\documentclass{article}

\usepackage[final,nohyperref]{corl_2022} % Uncomment for the camera-ready ``final'' version.
\usepackage{amsfonts}

\usepackage{color,xcolor}
\usepackage{epsfig}
\usepackage{graphicx}

\usepackage{adjustbox}
\usepackage{array}
\usepackage{booktabs}
\usepackage{colortbl}
\usepackage{float,wrapfig}
\usepackage{hhline}
\usepackage{multirow}
\usepackage{subcaption} % issues a warning with CVPR/ICCV format

\usepackage{amsmath,amsfonts,amsthm,amssymb}
\usepackage{pifont}% http://ctan.org/pkg/pifont
\usepackage{bm}
\usepackage{nicefrac}
\usepackage{microtype}

\usepackage{changepage}
\usepackage{extramarks}
\usepackage{fancyhdr}
\usepackage{lastpage}
\usepackage{setspace}
\usepackage{soul}
\usepackage{xspace}

\usepackage{url}

\usepackage{algorithm, algorithmic}
\usepackage{enumerate}
\usepackage{todonotes} % conflict with CVPR/ICCV format

\usepackage{titlesec}
\usepackage{makecell}

\usepackage{amsmath,amsthm,amssymb}
\usepackage{mathtools}
\usepackage{stmaryrd}
\usepackage{enumitem}

\usepackage{tabularx}

\newcolumntype{L}[1]{>{\raggedright\let\newline\\\arraybackslash\hspace{0pt}}m{#1}}
\newcolumntype{C}[1]{>{\centering\let\newline\\\arraybackslash\hspace{0pt}}m{#1}}
\newcolumntype{R}[1]{>{\raggedleft\let\newline\\\arraybackslash\hspace{0pt}}m{#1}}

\makeatletter
\DeclareRobustCommand\onedot{\futurelet\@let@token\@onedot}
\def\@onedot{\ifx\@let@token.\else.\null\fi\xspace}

\g@addto@macro\normalsize{%
  \setlength\abovedisplayskip{0pt}
  \setlength\belowdisplayskip{0pt}
  \setlength\abovedisplayshortskip{0pt}
  \setlength\belowdisplayshortskip{0pt}
}
\makeatother

\newcommand{\ourshort}{R-NDF\xspace}

\newcommand{\sect}[1]{Section~\ref{#1}}

\newcommand{\ignore}[1]{}

\newcommand{\myparagraph}[1]{\vspace{-5pt} \paragraph{#1}}

\newcommand{\coord}{\mathbf{x}}

\newcommand{\sethree}{$\text{SE}(3)$}

\newcommand{\encoder}{\mathcal{E}}
\newcommand{\sothree}{$\text{SO}(3)$}

\newcommand{\pointcloud}{\mathbf{P}}
\newcommand{\object}{\mathbf{O}}
\newcommand{\pose}{\mathbf{T}}
\newcommand{\pointndf}{f}
\newcommand{\posendf}{F}

\newcommand{\posedescriptor}{\mathcal{Z}}

\newcommand{\demopointcloud}{\hat{\pointcloud}}
\newcommand{\demoobject}{\hat{\object}}

\newcommand{\demopose}{\hat{\pose}}

\newcommand{\probes}{\mathcal{X}}
\newcommand{\rotaa}{\mathbf{R}_{\text{aa}}}

\DeclareMathOperator*{\argmin}{arg\,min}

\usepackage{color}
\definecolor{turquoise}{cmyk}{0.65,0,0.1,0.3}
\definecolor{purple}{rgb}{0.65,0,0.65}
\definecolor{dark_green}{rgb}{0, 0.5, 0}
\definecolor{orange}{rgb}{0.8, 0.6, 0.2}
\definecolor{red}{rgb}{0.8, 0.2, 0.2}
\definecolor{darkred}{rgb}{0.6, 0.1, 0.05}
\definecolor{blueish}{rgb}{0.0, 0.3, .6}
\definecolor{light_gray}{rgb}{0.7, 0.7, .7}
\definecolor{pink}{rgb}{1, 0, 1}
\definecolor{greyblue}{rgb}{0.25, 0.25, 1}

\definecolor{blueish}{rgb}{0.0, 0.3, .6}

\DeclareRobustCommand{\vsnote}[1]{\ifthenelse{\boolean{draft-mode}}{\textcolor{dark_green}{[VS: #1}]}{}}

\DeclareRobustCommand{\pulkit}[1]{\ifthenelse{\boolean{draft-mode}}{\textcolor{green}{\textbf{[PA: #1}]}}{}}
\DeclareRobustCommand{\asnote}[1]{\ifthenelse{\boolean{draft-mode}}{\textcolor{blue}{\textbf{AS: #1}}}{}}
\DeclareRobustCommand{\arnote}[1]{\ifthenelse{\boolean{draft-mode}}{\textcolor{green}{\textbf{AR: #1}}}{}}

\newcommand{\Figure}[1]{Fig.~\ref{fig:#1}}

\newcommand{\eq}[1]{(\ref{eq:#1})}

\newcommand{\Equation}[1]{Equation~\eq{#1}}

\newcommand{\Section}[1]{Section~\ref{sec:#1}}

\usepackage{blindtext}

\renewcommand{\paragraph}[1]{\vspace{.1em}\noindent\textbf{#1}.}

\usepackage{amsmath,amsfonts,bm}

\def\eqref#1{equation~\ref{#1}}
\def\1{\bm{1}}

\DeclareMathAlphabet{\mathsfit}{\encodingdefault}{\sfdefault}{m}{sl}
\SetMathAlphabet{\mathsfit}{bold}{\encodingdefault}{\sfdefault}{bx}{n}

\DeclarePairedDelimiterX{\infdivx}[2]{(}{)}{%
  #1\;\delimsize|\delimsize|\;#2%
}

\title{SE(3)-Equivariant Relational Rearrangement with Neural Descriptor Fields}

\author{
    Anthony Simeonov$^{*,1,2}$,
    Yilun Du$^{*,2}$,
    Lin Yen-Chen$^{2}$\\
    \textbf{Alberto Rodriguez}$^{3}$,
    \textbf{Leslie Pack Kaelbling}$^{2}$,
    \textbf{Tom\'as Lozano-P\'erez}$^{2}$,
    \textbf{Pulkit Agrawal}$^{1,2}$ \\
    Massachusetts Institute of Technology\\
    $^{1}$ Improbable AI Lab, $^{2}$CSAIL, $^{3}$Department of Mechanical Engineering, $^{*}$Equal Contribution
}

\newboolean{draft-mode}
\setboolean{draft-mode}{false}

\begin{document}
\maketitle   

%===============================================================================
\begin{abstract}
We present a method for performing tasks involving spatial relations between novel object instances initialized in arbitrary poses directly from point cloud observations. Our framework provides a scalable way for specifying new tasks using only $\sim$5-10 demonstrations. 
Object rearrangement is formalized as the question of finding actions that configure task-relevant parts of the object in a desired alignment. This formalism is implemented in three steps: assigning a consistent local coordinate frame to the task-relevant object parts, determining the location and orientation of this coordinate frame on unseen object instances, and executing an action that brings these frames into the desired alignment.
We overcome the key technical challenge of determining task-relevant local coordinate frames from a few demonstrations by developing an optimization method based on Neural Descriptor Fields (NDFs) and a \emph{single} annotated 3D keypoint.
An energy-based learning scheme to model the joint configuration of the objects that satisfies a desired relational task further improves performance. 
The method is tested on three multi-object rearrangement tasks in simulation and on a real robot.
Project website, videos, and code: \url{https://anthonysimeonov.github.io/r-ndf/}
\end{abstract}


%===============================================================================
\section{Introduction}
\label{sec: introduction}
% \vspace{-8pt}

\begin{figure*}[t]
\includegraphics[width=\linewidth]{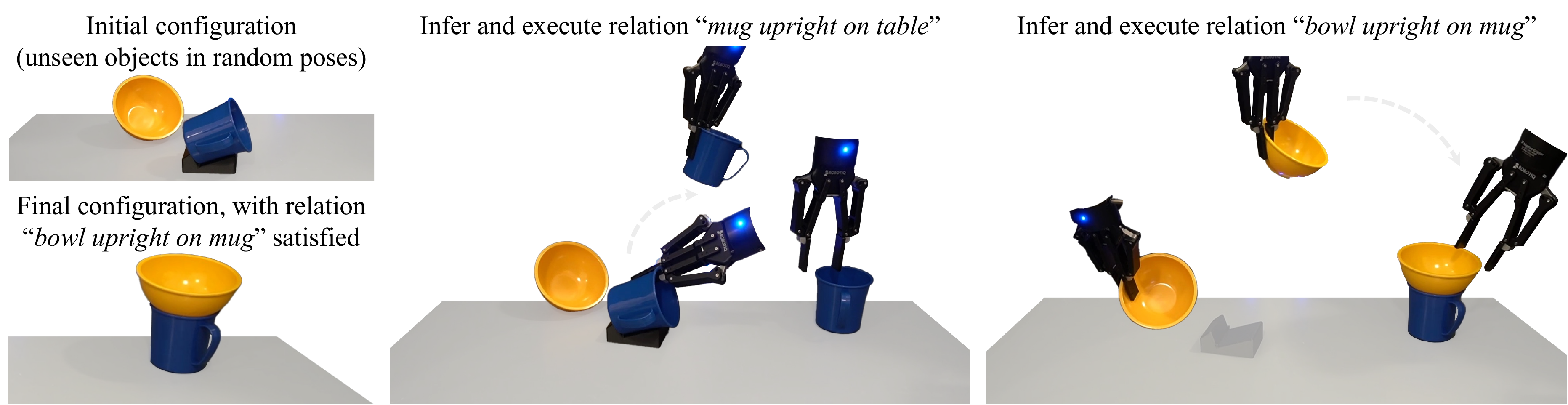}
\caption{ \small  Given a point cloud of a pair of unseen objects in arbitrary initial configurations (top left), Relational Neural Descriptor Fields (\ourshort{}s) obtain relative transformations that satisfy a relational task objective, such as ``placing the mug upright on the table'' (middle) and ``stacking the bowl upright on top of the mug'' (right). Our framework obtains these transformations by inferring the 6D pose of local coordinate frames at the task-relevant parts of the objects using just a small handful ($\sim$5-10) of demonstrations of each relational task.} 
\label{fig:intro-teaser}
\vspace{-15pt}
\end{figure*}

Many tasks we want robots to perform -- e.g., stacking bowls and plates to declutter a table, putting objects together to build an assembly, and hanging mugs on a rack with hooks -- involve rearranging objects relative to one another. 
Such tasks can be described in terms of spatial relations between \emph{parts} of a set of rigid objects. The desired relation can be achieved by first attaching a \emph{local} coordinate frame to \textit{task-relevant} parts of the object and then transforming the objects in a way that brings these coordinate frames into the desired alignment. 
For example, \emph{hanging} a mug on a rack is a relation between the mug's handle and the rack's hook, while \emph{stacking} a bowl on a mug involves aligning the bottom of the bowl with the top of the mug (see \Figure{intro-teaser}).

Specifying and solving tasks in this way requires the ability to (i) assign a \emph{consistent} local coordinate frame to the \emph{task-relevant} object parts, and (ii) \emph{detect} the corresponding coordinate frames on new object instances. % to then bring them into the desired alignment.
Some prior works use large task-specific datasets with human-labeled keypoints that identify the task-relevant parts~\cite{manuelli2019kpam, gao2021kpam}, but heavy dependence on manual annotation limits easy deployment of such approaches for a wide diversity of tasks. 
Neural Descriptor Fields (NDFs)~\cite{simeonov2022neural} overcome the need for large-scale annotation by leveraging task-agnostic self-supervised pretraining, followed by just a small set of task demonstrations ($\sim$ 5-10) to both identify the task-relevant object parts and assign each part an oriented local coordinate frame. 
NDFs have been shown to successfully localize these local coordinate frames at the corresponding parts of new object instances.

While NDFs require less task-specific data, labeling the relevant object parts in a consistent fashion can still be tedious -- e.g., one must assign an orientation to the ``handle'' of multiple mugs and ensure they are all consistent.
Prior work~\cite{simeonov2022neural} instead used demonstrations of the \emph{relation} to associate a \emph{single} frame, assigned to the \emph{second} object, with the task-relevant part of each manipulated object (e.g., label a frame on the ``hook'' of a rack \emph{once}, and associate this frame with each mug's ``handle'' based on the demonstrated interaction between the ``handle'' and the ``hook'').
However, this makes the limiting assumption that the secondary object is \emph{known}~\cite{simeonov2022neural} -- in the \emph{hanging} example, the system generalizes to unseen mugs, but fails if the rack is in a new pose or has a different shape.
Our work addresses this fundamental limitation of using NDFs for relational tasks. 
We present Relational Neural Descriptor Fields ({\ourshort}s), a framework, using $\sim$ 5-10 demonstrations, that takes as input 3D point clouds of a \emph{pair} of unseen objects in arbitrary initial poses and outputs a relative transformation between them that satisfies a relational task objective.

The central difficulty in applying NDFs to scenarios with changing pairs of objects is to assign a set of \emph{consistent} local coordinate frames to the \emph{task-relevant parts} of the objects in the demonstrations, which may be both \emph{unaligned} and \emph{differently shaped}. 
We propose an optimization method that uses two NDFs (one per object) and a \emph{single} 3D keypoint label in just \emph{one} of the demonstrations, to assign a set of local coordinate frames that are consistently posed relative to the task-relevant parts of the objects. 
We then apply NDFs to localize the corresponding coordinate frames for unseen pairs of objects presented in arbitrary initial poses, and solve for the relative transformation between them that satisfies the desired relation.
However, errors can accumulate when inferring a relative transformation based on a pair of coordinate frames that have  been \emph{independently} localized.
To mitigate this effect, we also propose a learning approach that directly models the \emph{joint} configuration of the pair of objects and helps refine the transformation for satisfying the relation.
We validate {\ourshort}s on three relational rearrangement tasks in both simulation and the real world.
Our simulation results show that \ourshort{}s outperform a set of baseline approaches, and our proposed optimization and learning-based refinement schemes benefit overall task success. %are advantageous.
Finally, our real world results exhibit the effectiveness of \ourshort{}s on pairs of diverse real world objects in tabletop pick-and-place, and highlight the potential for applying our approach to multi-step tasks.

%===============================================================================
% \vspace{-5pt}
\vspace{3pt}
\section{Background: Neural Descriptor Fields}
\vspace{2pt}
\label{sec:method-ndf}
% \vspace{-5pt}

A Neural Descriptor Field (NDF) \citep{simeonov2022neural} represents an object using a function $\pointndf$ that maps a 3D coordinate $\coord \in \mathbb{R}^{3}$ and an object point cloud $\pointcloud \in \mathbb{R}^{3\times N}$ to a spatial descriptor in $\mathbb{R}^{d}$: 
\begin{equation}
    \pointndf(\coord | \pointcloud): \mathbb{R}^3 \times \mathbb{R}^{3 \times N} \to \mathbb{R}^d.
\end{equation}
The function $\pointndf$ is parameterized as a neural network constructed to be \sethree{}-equivariant, such that if an object is subject to a rigid body transform $\pose \in$ \sethree{} its spatial descriptors transform accordingly\footnote{\vspace{0pt}We use homogeneous coordinates for ease of notation, i.e., $\pose \coord$ denotes $\mathbf{R}\coord + \mathbf{t}$ where $\pose = (\mathbf{R}, \mathbf{t}) \in~\sethree{}$.}:
%
% \vspace{-10pt}
% \vspace{1pt}
% \vspace{0.5pt}
\begin{align}
    \pointndf(\coord | \pointcloud) \equiv \pointndf(\pose \coord | \pose \pointcloud ).
\end{align}

This enables NDFs to behave consistently for the same object, regardless of the underlying pose.
NDFs are also trained to learn correspondence over objects in the same category, so that points near similar geometric features of different instances (e.g., a point near the handle of two different mugs) are mapped to similar descriptor values.
The equivariance property is obtained by using SO(3)-equivariant neural network layers~\cite{deng2021vector} and mean-centered point clouds, while the category-level correspondence is obtained by training $\pointndf$ on a category-level 3D reconstruction task~\cite{simeonov2022neural, mescheder2019occupancy}.

NDFs can also be redefined to model a field over full \sethree{} poses, rather than individual points.
This is achieved by concatenating the descriptors of the individual points in a \emph{rigid set} of query points $\probes \in \mathbb{R}^{3\times N_q}$, i.e., a set of three or more non-collinear points $\coord_i, ~i=1...N_q$, that are constrained to transform together rigidly.
This construction allows NDFs to represent an \sethree{} pose $\pose$ via its action on $\probes$, i.e., via the coordinates of the \emph{transformed query point cloud} $\pose \probes$: %(where $\probes^h$ denotes $\probes$ expressed with homogeneous coordinates):
\vspace{2pt}
\begin{equation}
\posedescriptor = \posendf(\pose | \pointcloud) = \bigoplus_{\coord_i \in \probes} \pointndf(\pose \coord_i |\pointcloud)
\label{eq:pose_enc}
\end{equation}
\vspace{2pt}
Thus, $\posendf$ maps a point cloud $\pointcloud$ and an \sethree{} pose $\pose$ to a category-level pose descriptor $\posedescriptor \in \mathbb{R}^{d \times N_{q}}$, where $\posendf$ inherits the same \sethree{}-equivariance from $\pointndf$.

% \vspace{-10pt}
\vspace{-3pt}
\section{General Problem Setup and Preliminaries}
\vspace{-2pt}
% \vspace{-5pt}
\label{sec:problem-setup}

Our high-level goal is to enable a user to specify a task involving a geometric relationship between a pair of rigid objects, and enable a robot to perform this task on unseen object instances presented in arbitrary initial poses. 
Examples of relations we consider include ``\emph{mug hanging on a rack}'', ``\emph{bowl stacked upright on a mug}'', and ``\emph{bottle placed upright on a tray}''.

Concretely, our goal is to build a system that takes as input two (nearly complete) 3D point clouds $\pointcloud_{A}$ and $\pointcloud_{B}$ (each segmented out from the overall scene) of objects $\object_{A}$ and $\object_{B}$, and outputs an \sethree{} transformation $\pose_{B}$ for transforming $\object_{B}$ into a configuration that satisfies a desired relation between $\object_{A}$ and $\object_{B}$. 
We represent the relation as an alignment between a pair of local coordinate frames attached to task-relevant geometric features of the objects, and break down the problem of obtaining $\pose_{B}$ into (i) assigning a set of consistent coordinate frames to the task-relevant local object parts and (ii) localizing these coordinate frames on the relevant parts of the new objects.

Furthermore, we assume a user specifies the relational task by providing a small handful of $K$ task demonstrations $\{\mathcal{D}_{i}\}_{i=1}^{K}$, such that it's intuitive and efficient to specify a wide diversity of tasks with minimal engineering effort.
A demonstration $\mathcal{D}$ consists of point clouds $\demopointcloud_{A}$ and $\demopointcloud_{B}$ (of objects $\demoobject_{A}$ and $\demoobject_{B}$) and relation-satisfying transformation $\demopose_{B}$. 

\paragraph{NDFs for Encoding Single Unknown Object Relations} 
Prior work on NDFs may be applied to a simplified version of this task, where the geometry and state of $\object_{A}$ is \emph{known}.
Given that $\object_{A}$ is known, we can initialize a set of query points $\probes_{A}$ near the task-relevant part of $\object_{A}$ and use the query points to encode the relative pose $\demopose_{B}$ via \Equation{pose_enc}. 
Thus, a demonstration $\mathcal{D}$ is mapped to a target pose descriptor $\hat{\posedescriptor} = \posendf(\demopose_{B}^{-1} | \demopointcloud_{B})$ representing the (inverse of the) final pose of $\demoobject_{B}$ \emph{relative to $\object_{A}$}. In practice, pose descriptors from multiple demonstrations $\{\mathcal{D}_{i}\}_{i=1}^{K}$ are averaged to obtain an overall descriptor $\hat{\posedescriptor} = \frac{1}{K} \sum_{i=1}^{K} \hat{\posedescriptor}_{i}$ for the whole set, which has important implications in the version of the task with two unknown objects (see \Section{multi-ndf} for further discussion). 

Given a novel object instance represented by point cloud~$\pointcloud_{B}$, we can compute a transformation $\pose_{B}$ such that transforming $\object_{B}$ by $\pose_{B}$ satisfies the demonstrated relation between $\object_{A}$ and $\object_{B}$. This is achieved by minimizing the L1 distance to the target pose descriptor $\hat{\posedescriptor}$:
\vspace{2pt}
\begin{equation}
\pose_{B}^{-1} = \underset{\pose}{\text{argmin}} \|
 \posendf(\pose | \pointcloud_{B}) - \hat{\posedescriptor}  \|.
\label{eq:energy}
\end{equation}
\vspace{2pt}
Intuitively, \Equation{energy} performs well across different objects due to the fact that NDFs are pretrained to enable reconstruction across a large dataset of 3D shapes.
As a result, shared descriptors are discovered across different instances in a shape category.
In contrast, training a model directly on the few demonstrations (e.g., for regressing pose $\pose_{B}$) would be more susceptible to overfitting.

% \vspace{-10 pt}
\section{Method}
% \vspace{-5pt}
\label{sec:method}

We now describe how we apply NDFs to infer relations between pairs of unknown objects. In \sect{sec:multi-ndf}, we propose an iterative optimization method for assigning consistent task-relevant coordinate frames to multiple objects. In \sect{sec:ebm-relations}, we discuss how we train a neural network on top of NDF features to model the joint object configuration and refine an inferred transformation. 
The system inputs consist of pretrained NDFs $\pointndf_{A}$ and $\pointndf_{B}$ for each object category, demonstrations $\{\mathcal{D}_{i}\}_{i=1}^{K} = \{(\demopointcloud_{A}, \demopointcloud_{B}, \demopose_{B})_{i}\}_{i=1}^{K}$, and a \emph{single} labeled 3D coordinate $\coord_{AB}$ for \emph{one} of the demonstrations, indicating approximately where the respective demonstration objects interact.

% \vspace{-3pt}
\subsection{Multiple NDFs for Inferring Pairs of Task-Relevant Local Coordinate Frames}
% \vspace{-3pt}
\label{sec:multi-ndf}

\begin{figure*}[t]
\includegraphics[width=\linewidth]{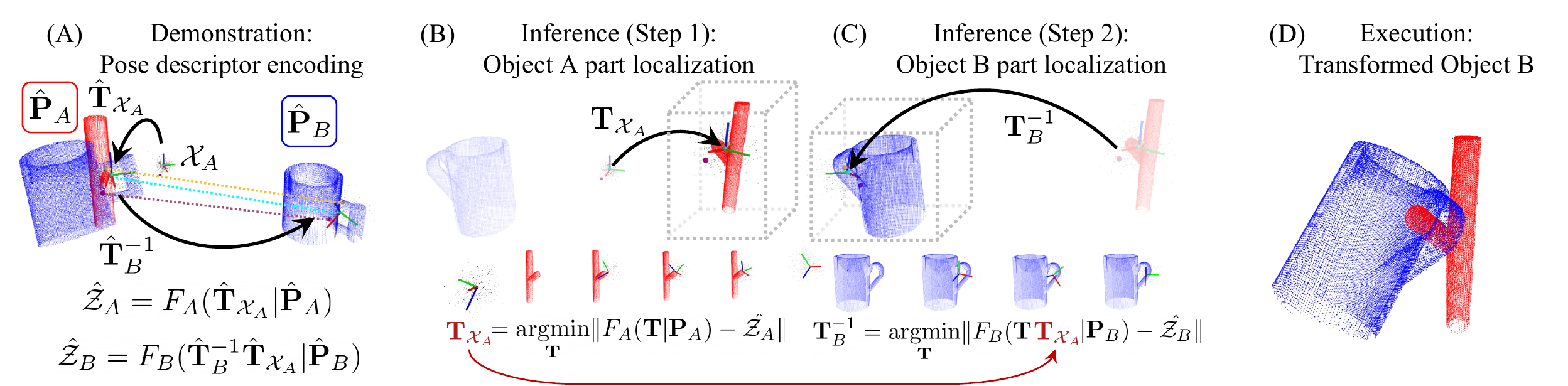}
\caption{ \small \textbf{Method Overview.} \textbf{(A)} A demonstration ($\demopointcloud_{A}, \demopointcloud_{B}, \demopose_{B}$) of a relation is encoded into a pair of pose descriptors by randomly sampling a set of query points $\probes_{A}$ at the origin and transforming it by $\demopose_{\probes_{A}}$ to be near the task-relevant interaction point $\coord_{AB}$. NDFs $\pointndf_{A}$ and $\pointndf_{B}$ are then used to obtain descriptors $\hat{\posedescriptor}_{A}$ and $\hat{\posedescriptor}_{B}$ representing coordinate frames near the task-relevant local parts on the objects. \textbf{(B)} Given point cloud $\pointcloud_{A}$ of a novel object, NDF $\pointndf_{A}$, and pose descriptor $\hat{\posedescriptor}_{A}$, pose $\pose_{\probes_{A}}$ of the corresponding coordinate frame on $\pointcloud_{A}$ is found. \textbf{(C)}. This procedure is then repeated with $\pointcloud_{B}$, $\pointndf_{B}$, and $\hat{\posedescriptor}_{B}$ to find pose $\pose_{B}^{-1}$ of the relevant parts of $\pointcloud_{B}$, \emph{relative to pose} $\pose_{\probes_{A}}$ \emph{found in the first inference step}. \textbf{(D)} Transforming $\pointcloud_{B}$ by $\pose_{B}$ satisfies the desired relation.}
\label{fig:demo-inference-exec}
% \vspace{-45pt}
% \vspace{-20pt}
\end{figure*}

Consider a scenario where $\object_{A}$ and $\object_{B}$ have \emph{unknown} underlying shapes and configurations.
We now show how NDFs can be used for inferring a \emph{pair} of task-relevant local coordinate frames on both objects and recovering a transformation $\pose_{B}$ that satisfies the relation.
The key idea of our approach is to formulate this problem as a bi-level optimization (illustrated in Figure \ref{fig:demo-inference-exec}),  where we first optimize to find a task-relevant portion of $\object_{A}$, and subsequently optimize a relative transform of a local part of $\object_{B}$ with respect to the local region of $\object_{A}$. % % from a set of demonstrations that contain a diverse set of parent objects in varying poses and 

We begin with two pretrained NDFs, $\pointndf_{A}$ and $\pointndf_{B}$, and query points $\probes_{A}$ in a canonical pose at the world frame origin. We obtain $\probes_{A}$ by sampling $N_q$ points from a zero-mean Gaussian and scaling such that $\probes_{A}$ has scale similar to the salient object parts. 
We then use the keypoint $\coord_{AB}$ to transform $\probes_{A}$ near the task-relevant features in the demonstration associated with $\coord_{AB}$. 
Denote this transformation as $\demopose_{\probes_{A}}$.
Finally, we encode \emph{world-frame} pose $\demopose_{\probes_{A}}$ into a descriptor conditioned on $\demopointcloud_{A}$, as $\hat{\posedescriptor}_{A} = \posendf_{A}(\demopose_{\probes_{A}} | \demopointcloud_{A})$, and \emph{relative} pose $\demopose_{B}^{-1}$ as $\hat{\posedescriptor}_{B} = \posendf_{B}(\demopose_{B}^{-1} \demopose_{\probes_{A}} |  \demopointcloud_{B})$, conditioned on $\demopointcloud_{B}$.
At test-time, we optimize both the world-frame pose of the query points $\pose_{\probes_{A}}$ and the (inverse of) pose $\pose_{B}$ relative to the initial pose found in the first step:
\bgroup
\def\arraystretch{0.4}
\noindent\begin{tabularx}{\textwidth}{@{}XX@{}}
  \begin{equation}
  \pose_{\probes_{A}} = \underset{\pose}{\text{argmin}} \|
 \posendf_{A}(\pose | \pointcloud_{A}) - \hat{\posedescriptor}_{A}  \|
\label{eq:multi-energy-a}
  \end{equation} &
  \begin{equation}
  \pose_{B}^{-1} = \underset{\pose}{\text{argmin}} \|
 \posendf_{B}(\pose \pose_{\probes_{A}} | \pointcloud_{B}) - \hat{\posedescriptor}_{B}  \|
    \label{eq:multi-energy-b}
  \end{equation}
\end{tabularx}
\egroup
Figure~\ref{fig:demo-inference-exec} shows an example of this pipeline, where the resulting $\pose_{B}$ is applied to the point cloud $\pointcloud_{B}$ of $\object_{B}$ to satisfy the ``hanging'' relation. 

\begin{wrapfigure}{r}{.4\linewidth}
\vspace{-15pt}
\centering
\includegraphics[width=\linewidth]{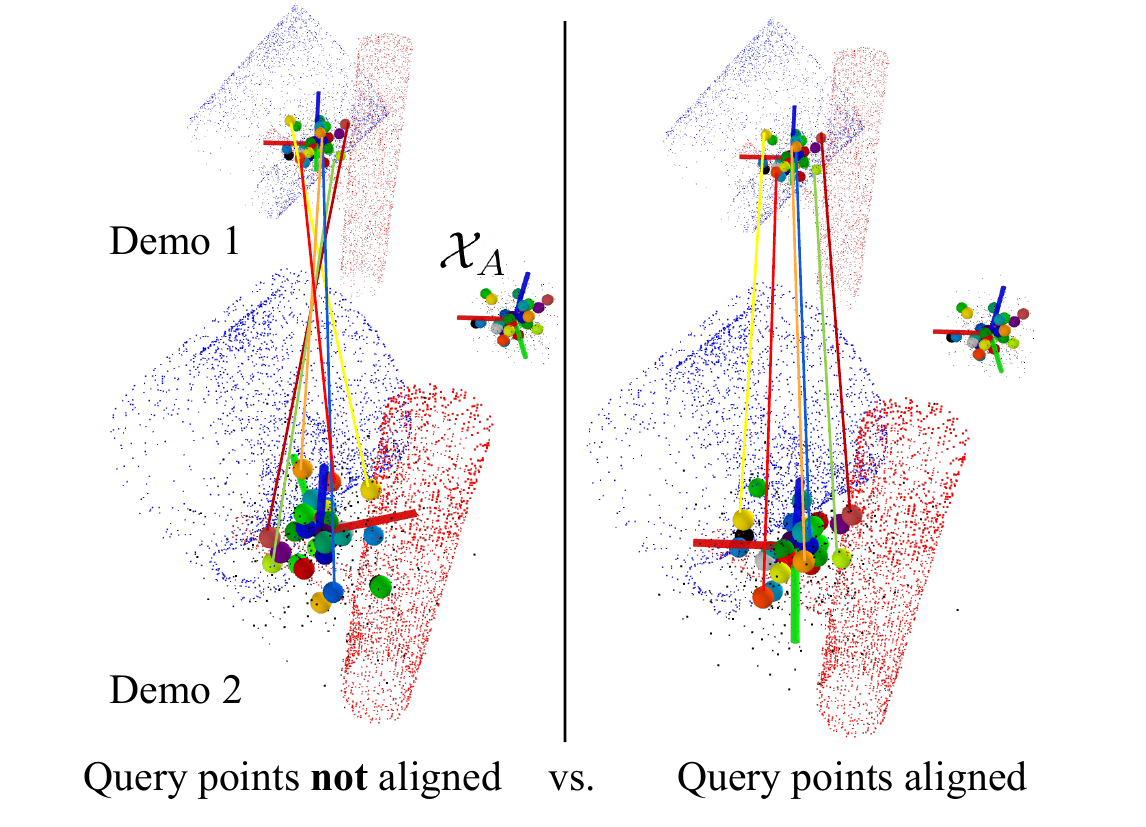}
\caption{ \small \textbf{Demo alignment.} We align the query points by minimizing the variance across the descriptor set before averaging.}
\vspace{-10pt}
\label{fig:aligned_vs_unaligned}
\end{wrapfigure}

\paragraph{Minimizing Descriptor Variance} 
In practice, solving Equations (\ref{eq:multi-energy-a}) and (\ref{eq:multi-energy-b}) works better if pose descriptors $\{\hat{\posedescriptor}_{i}\}_{i=1}^{K}$ from multiple demonstrations are averaged together to obtain an overall target descriptor $\hat{\posedescriptor} = \frac{1}{K} \sum_{i=1}^{K} \hat{\posedescriptor}_{i}$ (see Sec. \ref{sec:results-sim} and~\cite{simeonov2022neural}). 
The reason is that a single demonstration \emph{underspecifies} which object parts are relevant for the task, allowing $\hat{\posedescriptor}$ to be sensitive to object features that are not relevant to the desired relation.
Instead, a set of demonstrations using slightly different objects (e.g., with different scales) reveals regions near local interactions that are \emph{shared} across the demonstrations, which helps disambiguate between parts that are critical vs. irrelevant for the specified relation. 
However, to avoid the pitfalls of averaging across a potentially multimodal or disjoint set, we want descriptors in the set $\{\hat{\posedescriptor}_{i}\}_{i=1}^{K}$ to be sensitive to nearby local geometry in a way that is \emph{consistent} (i.e., \emph{unimodal}) across the demos.
This only occurs if the \emph{query points} used to obtain the descriptors are themselves consistently aligned relative to each respective object (see Figure~\ref{fig:aligned_vs_unaligned}).
Therefore, we need to find a transformation $\demopose_{\probes_{A}, i}$ for \emph{each} demonstration $\mathcal{D}_{i}$ that transforms the canonical query points $\probes_{A}$ into a configuration that leads the descriptors $\{\hat{\posedescriptor}_{i} = \posendf_{A}(\demopose_{\probes_{A}, i} | \demopointcloud_{A, i})\}_{i=1}^{K}$ to be consistent with each other. 
We address this by finding the set of transformations $\{\demopose_{\probes_{A}, i}\}_{i=1}^{K}$ that minimizes the \emph{variance} across the descriptor set $\{\hat{\posedescriptor}_{i} = \posendf_{A}(\demopose_{\probes_{A}, i} | \demopointcloud_{A, i})\}_{i=1}^{K}$:
\vspace{2pt}
\begin{align}
\label{eq:min-var}
\underset{\{\demopose_{\probes_{A}, i}\}_{i=1}^{K}}{\text{min}}~~ \text{Var}(\{ \hat{\posedescriptor}_{i} \}_{i=1}^{K}) \quad\text{subject to}~~\hat{\posedescriptor}_{i} = \posendf_{A}(\demopose_{\probes_{A}, i} | \demopointcloud_{A, i}) \quad \text{for } i = 1,...,K
\end{align}
where $\text{Var}(\cdot)$ denotes the sum of the per-element variance across a set of vectors.
We perform this minimization by applying NDFs in an alternating optimization procedure. Starting with an initial reference pose (constructed using $\coord_{AB}$) placing $\probes_{A}$ near the task-relevant object parts in one of the demonstrations, we iteratively apply \Equation{multi-energy-a} to obtain a descriptor for each demonstration that matches the reference. At the outer level, we refit the reference descriptor using the mean of the most recently obtained individual descriptors, and repeat.
More details can be found in the Appendix.

% \vspace{-5pt}
\subsection{Capturing Joint Descriptor Alignment through Learned Energy Functions}
\label{sec:ebm-relations}
The method in \sect{sec:multi-ndf} proposes to infer a desired relation by \emph{sequentially} localizing \emph{independent} coordinate frames for each object. While this approach is generally effective, small errors can accumulate and cause slight misalignments that lead to failure in the execution. We thus propose to learn a neural network which directly captures the \emph{joint} configuration of $\object_{A}$ and $\object_{B}$ that satisfies the desired relation, and use this model to refine predictions made by the method in \sect{sec:multi-ndf}.

\paragraph{Pairwise Energy Functions} We train an Energy-Based Model (EBM) $E_\theta(\cdot)$ \citep{Du2019ImplicitGA} to parameterize a learned energy landscape over NDF encodings of relative poses between $\object_{A}$ and $\object_{B}$ (i.e,. $E_{\theta}$ acts as a learned analogue for the L1 distance in \sect{sec:multi-ndf}). 
The energy function $E_\theta(\cdot)$ is trained so that the ground truth transform of $\object_{B}$ with respect to $\object_{A}$ is recovered given NDFs $\pointndf_{A}$ and $\pointndf_{B}$ (note that $\pointndf$ corresponds to descriptor evaluation at \textit{single} coordinate $\coord$ while $\posendf$ is defined over \textit{sets} of coordinates). Explicitly, our energy function is trained so that: 
\begin{equation}
\pose_{B} = \underset{\pose}{\text{argmin}}  \left [ 
   E_\theta(\pointndf_B(\cdot|\pose \pointcloud_{B}), \pointndf_A(\cdot| \pointcloud_{A})) \right].
\label{eq:energy_learned}
\end{equation}
Since each NDF is a continuous field, it is difficult to input them directly into our energy function $E_\theta(\cdot)$. We represent the energy function as the sum of the point-wise evaluation of each NDF on a set of different query points $\probes_{E}$ sampled from transformed pointcloud $\pose \pointcloud_B$.
\vspace{2pt}
\begin{equation}
   E_\theta(\pointndf_B(\cdot|\pose \pointcloud_B), \pointndf_A(\cdot| \pointcloud_A)) = \sum_{\coord \in \probes_{E}}  E_\theta(\pointndf_B(\coord|\pose \pointcloud_B), \pointndf_A(\coord|\pointcloud_A))
\end{equation}
\vspace{2pt}
At test-time, we use Equation~(\ref{eq:energy_learned}) to refine the transformation obtained using Equations (\ref{eq:multi-energy-a}) and (\ref{eq:multi-energy-b}).

% \vspace{-7pt}
\subsection{Learning}
\label{sec: implementation-training}

\paragraph{NDF training}
We represent NDFs $\pointndf_{A}$ and $\pointndf_{B}$ as two neural networks with identical architecture and separate weights. 
Following~\cite{simeonov2022neural}, the architecture consists of a PointNet~\cite{qi2017pointnet} point cloud encoder with SO(3) equivariant Vector Neuron~\cite{deng2021vector} layers, and a multi-layer perceptron (MLP) decoder. The NDF is represented as a function mapping a 3D coordinate and a point cloud to the vector of concatenated activations of the MLP. The models are trained end-to-end to reconstruct 3D shapes given object point clouds.
We use a dataset of ground truth 3D shapes and generate a corresponding set of 3D point clouds in simulation. More architecture and training data details can be found in the Appendix.

\paragraph{Energy-Based Model Training} 
We supervise the EBM $E_\theta$ so that optimization over the learned energy landscape recovers the relative transform between $\object_A$ and $\object_B$. In particular, we follow the training objective in \citep{Du2021UnsupervisedLO} and train $\argmin_{\pose}[E_\theta(\cdot)]$ to match a target pose using the following procedure. % $\pose^{*}$. 
We first apply a small delta perturbation $\pose_{\Delta}$ to $\demopose_{B}\demopointcloud_{B}$ (i.e., the point cloud of $\object_{B}$ in its final configuration) to obtain $\demopointcloud_{B, \Delta} = \pose_{\Delta}\demopose_{B}\demopointcloud_{B}$.
We then train $E_\theta$ to iteratively refine an initial random pose $\pose_{0}$ with translation $\mathbf{t}_0$ and rotation $\mathbf{R}_0$ to \emph{undo} the perturbation pose $\pose_{\Delta}$.
We run $n$ steps of optimization on $\mathbf{t}_0$ and $\mathbf{R}_0$, where an individual step is given by $\mathbf{t}_k = \mathbf{t}_{k-1} - \lambda \nabla_{\mathbf{t}} E_\theta(\pointndf_A(\cdot| \demopointcloud_A), \pointndf_B(\cdot|\pose \demopointcloud_{B, \Delta}))$ and $\mathbf{R}_k = \mathbf{R}_{k-1} - \lambda \nabla_{\mathbf{R}} E_\theta(\pointndf_c(\cdot| \demopointcloud_A), \pointndf_B(\cdot|\pose \demopointcloud_{B, \Delta}))$.
We may train the energy function so that $\pose_{n}$ corresponds to the inverse of the perturbation pose $\pose_{\Delta}$ using $\mathcal{L}_{\text{trans}} = \|\mathbf{t}_n - \mathbf{t}^{-1}_{\Delta}\|$  and $\mathcal{L}_{\text{rot}} = \|\mathbf{R}_n - \mathbf{R}^{-1}_{\Delta}\|$.
However, with symmetric objects, there are multiple different rotations $\mathbf{R}_n$ which may satisfy the desired relation (e.g., a bowl is still ``on'' a mug, regardless of the angle about its radial axis). 
To account for these symmetries, we implicitly enforce consistency between an optimized transform $\pose_{n}$ and $\pose^{-1}_{\Delta}$ by enforcing that its application on $\demopointcloud_{B, \Delta}$ leads to a similar point cloud to $\demopose_{B}\demopointcloud_{B}$. We achieve this by minimizing the Chamfer loss ~\cite{barrow1977parametric} between the optimized transformed point cloud $\pose_{n} \demopointcloud_{B, \Delta}$ and the demonstration point cloud $\demopose_{B} \demopointcloud_{B}$.

% \vspace{-15pt}
\vspace{-10pt}
\section{Application to Tabletop Manipulation}
\vspace{-5pt}
\label{sec: apply-to-manip}
\paragraph{Robot and Environment Setup}
We apply the method in \Section{method} to the problem of tabletop object rearrangement using a Franka Panda robotic arm with a Robotiq 2F140 parallel jaw gripper. 
The arm is used to collect the demonstrations and to execute the inferred transformation at test-time. 
Our environment consists of the arm on a table with four calibrated depth cameras.

\paragraph{Providing and Encoding Demonstrations}
When collecting a demonstration, initial object point clouds $\demopointcloud_{A}$ and  $\demopointcloud_{B}$ of objects $\demoobject_{A}$ and $\demoobject_{B}$ are obtained by fusing a set of back projected depth images. 
The demonstrator moves the gripper to a pose $\demopose_{\text{grasp}}$, grasps $\demoobject_{B}$, and finally moves the gripper to a pose $\demopose_{\text{place}}$ that satisfies the desired relation between $\demoobject_{A}$ and $\demoobject_{B}$.
$\demopose_{B}$ is obtained as $\demopose_{\text{place}}\demopose_{\text{grasp}}^{-1}$. % %
In \emph{one} of the demonstrations, a 3D keypoint $\coord_{AB}$ is labeled near the parts of the objects that interact with each other by moving the gripper to this region and recording its position. 

\paragraph{Test-time Task Setup and Inference}
At test time, we are given point clouds $\pointcloud_{A}$ and $\pointcloud_{B}$ of new objects $\object_{A}$ and $\object_{B}$.
Equations $(\ref{eq:multi-energy-a}$), ($\ref{eq:multi-energy-b}$), and ($\ref{eq:energy_learned}$) are applied in sequence to obtain $\pose_{B}$. 
$\pose_{B}$ is applied to $\object_{B}$ by transforming an initial grasp pose $\pose_{\text{grasp}}$ (obtained using a separate grasp generation pipeline) by $\pose_{B}$ to obtain a placing pose $\pose_{\text{place}} = \pose_{B}\pose_{\text{grasp}}$, and off the shelf inverse kinematics and motion planning is used to reach $\pose_{\text{grasp}}$ and $\pose_{\text{place}}$. 
% 

%===============================================================================

%===============================================================================
% \vspace{-5pt}
\section{Experiments and Results}
\label{sec:experiments}
% \vspace{-5pt}
Our experiments are designed to evaluate \ourshort{}s in executing relational rearrangement tasks with unseen objects using only a few demonstrations. 
We seek to answer three questions: (1) How well do \ourshort{}s predict transformations that satisfy a relational task? (2) How important is each component in \ourshort{}s? (3) Can \ourshort{}s be used to perform multi-object pick-and-place tasks in the real world?

We also show additional results regarding (i) multi-step rearrangement via relation sequencing, (ii) composing multiple energy terms in the optimization to achieve collision avoidance and multi-object rearrangement, and (iii) applying \ourshort{} with partial point clouds in the Appendix.

\paragraph{Baselines}
\label{sec:experiments-baselines}
As existing rearrangement methods are not directly applicable with so few demonstrations, we compare with two constructed baselines. The first is to train an MLP to directly regress the relative transformation between objects (``Pose Regression''). The MLP takes as input the point cloud encodings obtained from the same PointNet~\cite{qi2017pointnet} encoder with Vector Neuron~\cite{deng2021vector} layers used in NDFs, and is trained directly on the demonstrations.
The second method is based on 3D point cloud registration (``Patch Match''). We use a state-of-the-art registration method~\cite{gao2019filterreg} to align the test-time shapes to the demonstration shapes and then compute the resulting relative transformation.

\paragraph{Task Setup and Evaluation Metrics}
We consider three relational rearrangement tasks for evaluation: (1) Hanging a mug on the hook of a rack, (2) Stacking a bowl upright on top of a mug, and (3) Placing a bottle upright inside of a box-shaped container.
We provide 10 demonstrations of each task and evaluate if each method, using the demonstrations, can infer a transformation that satisfies the desired relation for unseen pairs of object instances with randomly sampled poses.
Experiments are conducted in both the real world and in simulation using PyBullet~\cite{coumans2016pybullet}.
In simulation, the transformation obtained by each method is directly applied by resetting the simulator to the transformed object states. 
To quantify performance, we report the success rate over 100 trials, where we use the ground truth simulator state to compute success (objects must be in contact, have the correct relative orientation, and not interpenetrate).

% \vspace{-5pt}
\subsection{Simulation Results}
\label{sec:results-sim}
\begin{figure}
    \centering
    \setlength{\tabcolsep}{0.3pt}
    \begin{tabular}{cc}
    \begin{subfigure}[b]{0.7\linewidth}
    \centering 

    \footnotesize
    \setlength{\tabcolsep}{2.3pt}
    \centering
\begin{tabular}{@{}lcccccc@{}}
    \toprule
      & \multicolumn{2}{c}{\textbf{Bowl on Mug}} & \multicolumn{2}{c}{\textbf{Mug on Rack}} & \multicolumn{2}{c}{\textbf{Bottle in Container}}  \\
     \cmidrule(lr){2-3} \cmidrule(lr){4-5} \cmidrule(lr){6-7} 
     \textbf{Method} & Upright &  Arbitrary & Upright & Arbitrary &  Upright & Arbitrary \\
     \midrule
       Pose Regression & 35.0 & 6.0 & 13.0 & 10.0 & 37.9 & 12.0 \\
       Patch Match & 34.0 & 32.0 & 56.0 & 44.0 & 44.0 & 42.0 \\
      \ourshort & \textbf{74.0} & \textbf{70.0} & \textbf{84.0} & \textbf{75.0} & \textbf{80.0} & \textbf{75.0} \\
    \bottomrule
    
\end{tabular}
    \caption{}    
    \label{tbl:benchmark}
     \end{subfigure} &
    \begin{subfigure}[b]{0.3\linewidth}
        \centering 
        \includegraphics[width=0.99\linewidth]{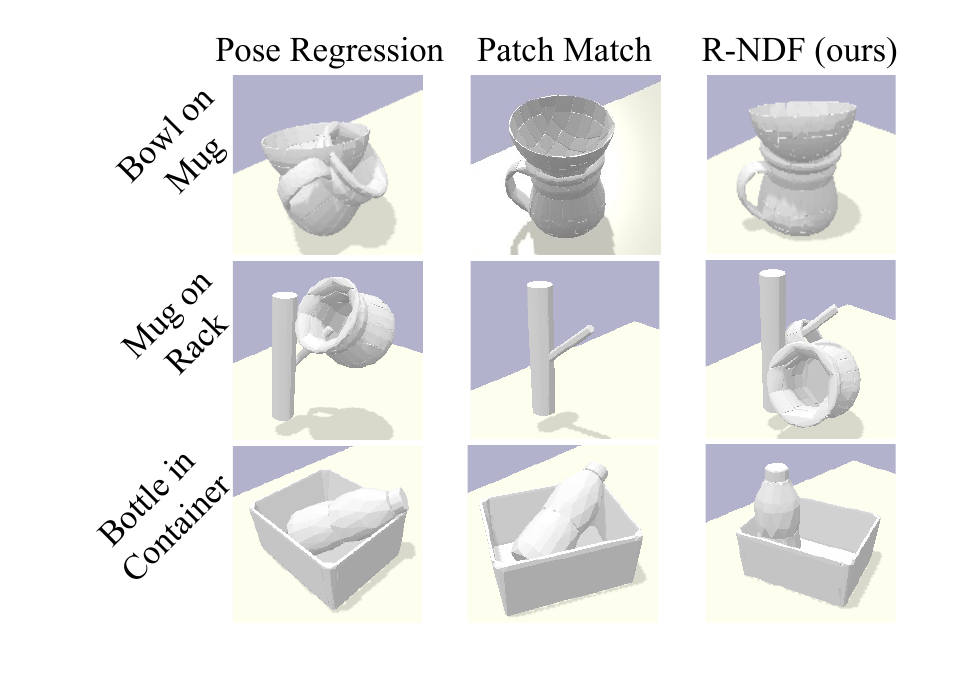}
        \vspace{-20pt}
 		\caption{}
    \label{fig:results-common-failures}
	\end{subfigure}
	
	\end{tabular}
	\vspace{-10pt}
     \caption{\small (a) \small \textbf{Relation inference success rates in simulation.} \ourshort performs better than the baseline approaches. (b) \textbf{Example predictions.} Representative predictions made by each method in simulation}
     \vspace{-10pt}
\end{figure}

% \vspace{-5pt}
%
We begin by evaluating how well \ourshort{}s can infer the desired transformations in simulation.
We consider two settings of varying difficulty.
First, the pair of unseen objects are positioned randomly on the table with a randomly sampled ``upright'' orientation (similar to those used in the demonstrations). Second, the orientation of $\object_{B}$ is randomly sampled from the full space of 3D rotations. 

Results in Table~\ref{tbl:benchmark} compare the performance of our approach to the baselines.
On the other hand, the registration-based method can sometimes find transformations that correctly align the unseen shapes to the demonstration objects, and thus achieves higher success rates than pose regression.
However, 3D registration is susceptible to locally optimal results that align the task \emph{irrelevant} parts of the objects. 
Common failure modes of using 3D registration in the tasks we consider include aligning the body of the mug but ignoring the handle, or aligning the racks to be upside down.
Figure~\ref{fig:results-common-failures} illustrates the final simulator state after applying some of the representative predictions of each method. 

In contrast, \ourshort{}s more accurately localize the task-relevant object parts and assign coordinate frames to these parts that are consistent with the demonstrations, leading to the highest success rates. 
Consistent with~\cite{simeonov2022neural}, the performance gap between the ``upright'' and ``arbitrary'' pose settings is small, which can be attributed to the built-in equivariance of the features used in \ourshort{}.

% \vspace{-5pt}
\subsection{Ablations}
% \vspace{-3pt}
\label{sec:experiments-ablations}

Next, we analyze the importance of the individual components of \ourshort{}s. 
We investigate ablations on the simulated ``mug on rack'' task, again considering both ``upright'' and ``arbitrary'' pose settings. 

\begin{figure}
    \centering
    \setlength{\tabcolsep}{0.3pt}

    \begin{tabular}{cc}
    \begin{subfigure}[b]{0.6\linewidth}
    \centering 

    \footnotesize
    \setlength{\tabcolsep}{2.3pt}
    \centering
    
    \begin{tabular}{ccc|cc}
        \toprule
        \textbf{Multiple}  & \textbf{Query Point} & \textbf{EBM} & \textbf{Upright} & \textbf{Arbitrary} \\
         \textbf{Demonstration} & \textbf{Alignment} & \textbf{Refinement} & \textbf{Pose} & \textbf{Pose} \\
        \midrule
        No & No & No & 39.3 & 43.6 \\
        Yes & No & No & 66.0 & 60.0\\
        Yes & Yes & No & 78.0 & 72.0 \\ 
        Yes & Yes & Yes & 84.0  & 75.0 \\
        \bottomrule
     \end{tabular}
    \caption{}    
     \label{tbl:ablation}
     \end{subfigure} &
    \begin{subfigure}[b]{0.4\linewidth}
        \centering 
		\includegraphics[width=1.0\linewidth]{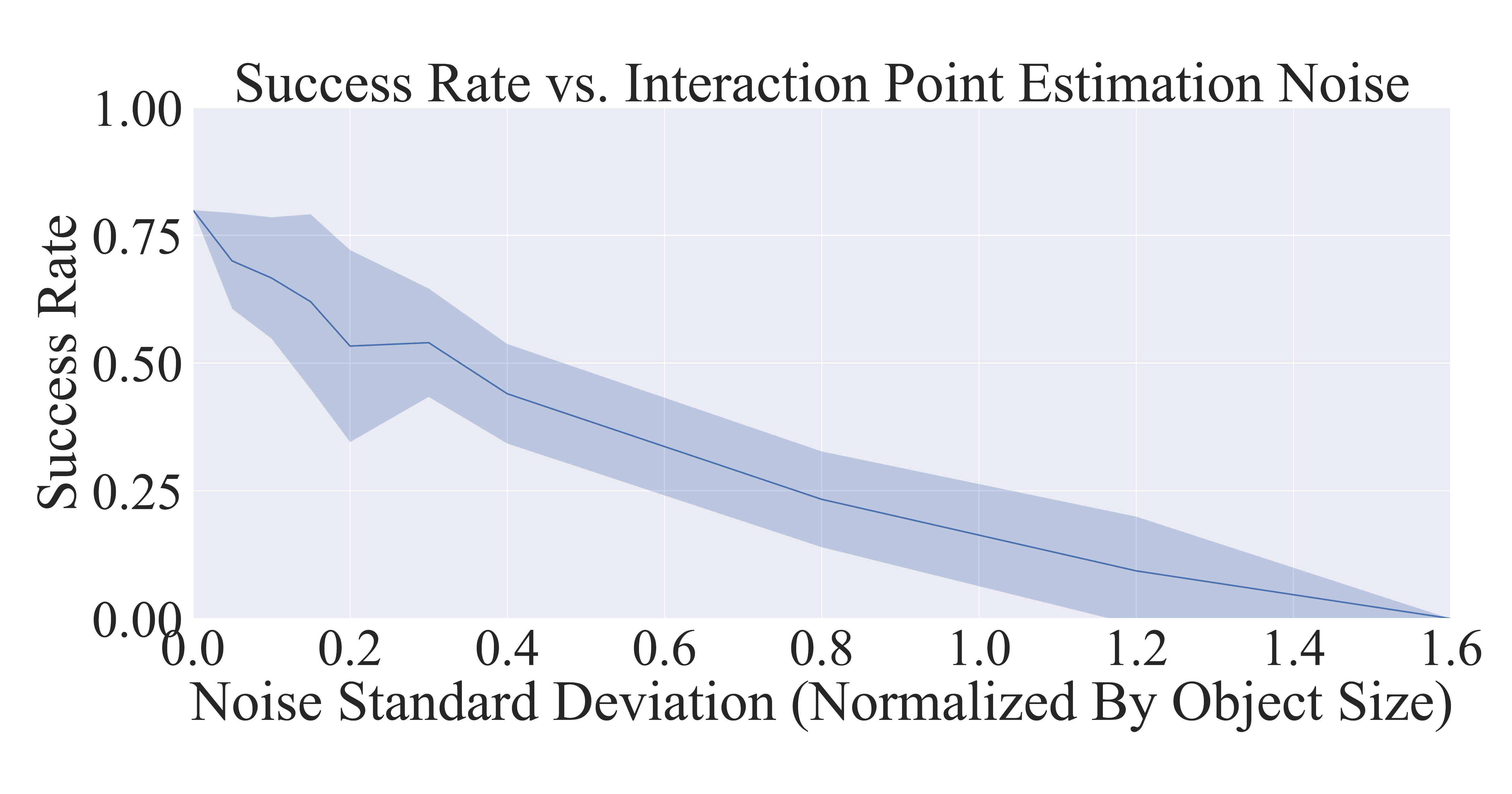}
        \vspace{-20pt}
 		\caption{}
		\label{fig:success-vs-noise}
	\end{subfigure}
	
	\end{tabular}
	\vspace{-10pt}
     \caption{ \small (a) \textbf{Ablations.} \ourshort performance with different components ablated. Success rate is highest when using multiple demonstrations, query point alignment, and EBM refinement. (b) \textbf{Success vs. Keypoint Noise.} Success rate vs. magnitude of noise (normalized by object size) added to the single labeled 3D keypoint $\coord_{AB}$.}
     \vspace{-15pt}
\end{figure}

The top row of Table~\ref{tbl:ablation} illustrates that \ourshort{} performs worse with a single demonstration. 
Since there are multiple possible explanations for the alignment between two objects when given one example of the desired relation, pose descriptors obtained from a single demonstration are more sensitive to task-irrelevant object features. 
The second row of Table~\ref{tbl:ablation} investigates the effect of averaging descriptors across the set of demonstrations without first aligning the query points relative to the objects in each demo.
We modified the demonstrations to provide keypoints $\{\coord_{AB, i}\}_{i=1}^{K}$ near the relevant region in \emph{each} demonstration, and then transform the query points to this region \emph{without} aligning their orientations.
Removing the query point alignment reduces the performance. 
The third row of Table~\ref{tbl:ablation} shows that removing the EBM refinement also decreases the success rate. 

We further examine the importance of accurately specifying the 3D keypoint $\coord_{AB}$ near the task-relevant region on one of the demonstrations.
We run the trials multiple times with Gaussian distributed noise added to the labeled point.
Figure~\ref{fig:success-vs-noise} shows a plot of the success rate vs. the noise magnitude normalized by the approximate size of the object.
The plot indicates that with limited noise perturbation, the success rate does not suffer significantly, though we observe a steep decline with more substantial perturbations.
These larger perturbations shift the query points to regions near geometric features that are less relevant to the desired relation.  

% \vspace{-5pt}
\subsection{Real Results}
% \vspace{-3pt}
\label{sec:results-real}

\begin{figure*}[t]
\includegraphics[width=\linewidth]{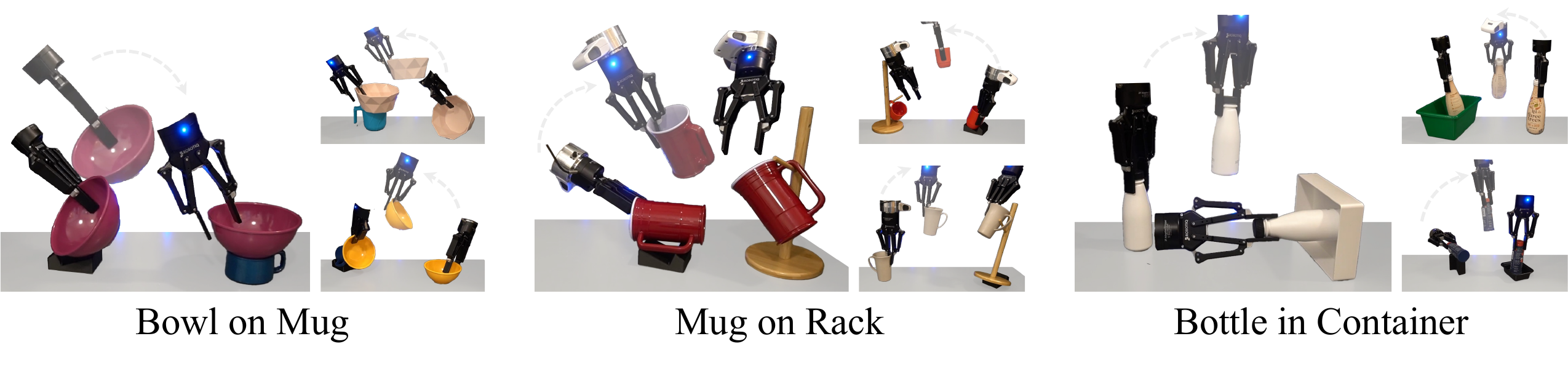}
\vspace{-15pt}
\caption{ \small \textbf{Real Execution Results.} Example executions of relational tasks on unseen mugs, bowls, bottles, racks, and containers in the real world. Our framework enables inferring the relative transformation between pairs of unseen objects in arbitrary initial poses from a small handful of unaligned demonstrations of each task.}
\label{fig:real-world-results}
\vspace{-15pt}
\end{figure*}

Finally, we validate that \ourshort{}s can be used to perform pick-and-place on pairs of unseen objects in the real world. 
Figure~\ref{fig:real-world-results} shows the execution on our three tasks. Our method successfully infers a transformation between the objects that satisfies the relations, despite the objects being presented in a challenging array of initial configurations. 
Figure~\ref{fig:intro-teaser} shows a multi-step rearrangement application of \ourshort{}s for the ``bowl on mug'' task. 
First, a relation between the \emph{mug} and the \emph{table} is specified and inferred for placing the mug upright. 
Then, the system executes the ``stacking'' relation between the \emph{bowl} and the \emph{upright mug}.
This highlights how \ourshort{}s can enable executing sequential chains of relations to satisfy task objectives involving more than two objects. Please see our attached supplemental video for additional real-world results.

%===============================================================================
% \vspace{-5pt}
\section{Related Work}
\label{sec: related}
% \vspace{-3pt}

\paragraph{Novel Object Rearrangement}
Several methods exist for novel object rearrangement~\cite{manuelli2019kpam,cheng2021learning_regrasp_place,li2022stable, thompson2021shape,batra2020rearrangement,simeonov2020long,lu2022online,gualtieri2021robotic, florence2019self, curtis2022long,paxton2022predicting,yuan2021sornet,goyal2022ifor,qureshi2021nerp,driess2021learning_ijrr,driess2020deep,liu2022structformer, goodwin2022semantically, danielczuk2021object}, many of which don't consider multiple varying objects that interact. 
CatBC~\cite{wen2022you} uses dense correspondence models to achieve impressive pick-and-place policy generalization from a single demonstration but assumes a known receptacle for placing.
Neural shape mating~\cite{chen2022shape_mating}, OmniHang~\cite{you2021omnihang}, and kPAM 2.0~\cite{gao2021kpam} generalize to pairs of unseen objects, but these approaches train on large task-specific datasets.
TransporterNets~\cite{zeng2020transporter,huang2022equivariant} enables rearrangement with varying pick and place locations from a few demonstrations, but focuses on top-down manipulation and struggles with out-of-plane reorientation. In contrast, we focus on executing relations involving large 3D reorientations.

\paragraph{Neural Fields in Robotics} 
Neural fields use neural networks to parameterize functions over continuous spatial or temporal coordinates~\cite{neuralfieldsvisualcomputing2022}.
They have been applied to model various signals and scene properties, such as images~\cite{karras2021alias}, geometry~\cite{park2019deepsdf,mescheder2019occupancy, chen2019learning}, appearance~\cite{mildenhall2020nerf,sitzmann2019scene}, tactile imprints~\cite{gao2021objectfolder}, and sound~\cite{luo2022learning}, with high fidelity and memory efficiency. Neural fields have been applied to represent objects for manipulation ~\cite{simeonov2022neural,ha2022deep,wi2022virdo,jiang2021synergies,jiang2022ditto} and environment states for dynamics and policy learning ~\cite{li2021visuomotor,driess2022learning,driess2022reinforcement}. They have also been used for pose estimation~\cite{yen2020inerf,adamkiewicz2022vision}, SLAM~\cite{moreau2022lens, Sucar:etal:ICCV2021}, and representing object geometry without depth cameras \cite{yen2022nerfsupervision,IchnowskiAvigal2021DexNeRF}. 
% \vspace{-5pt}
\section{Limitations and Conclusion}

\paragraph{Limitations} \ourshort{}s require a pretrained NDF for each category used in the task, which can be nontrivial to obtain for novel object categories without existing 3D model datasets.
Our approach also requires an annotated keypoint to localize task-relevant object parts.
Future work could explore automated discovery of task-relevant regions directly from a set of demonstrations.
Our system uses depth cameras, which often struggle with noise and objects with thin and transparent features.
An RGB-only approach offering a similar level of generalization would be interesting to investigate.
Finally, we require segmented object point clouds.
While object instance segmentation is quite mature, pretrained segmentation models regularly struggle when objects are in diverse orientations.  

\paragraph{Conclusion} This work presents an approach for learning from a limited number of demonstrations to rearrange novel objects into configurations satisfying a relational task objective. 
We develop methods that build upon prior applications of neural fields for representing objects and increase the scope of tasks they can achieve.
Our results illustrate the general applicability of our framework across a diverse range of relational tasks involving pairs of novel objects in arbitrary initial poses. 
%===============================================================================

% The maximum paper length is 8 pages excluding references and acknowledgements, and 10 pages including references and acknowledgements

\clearpage
% The acknowledgments are automatically included only in the final version of the paper.

% \acknowledgments{If a paper is accepted, the final camera-ready version will (and probably should) include acknowledgments. All acknowledgments go at the end of the paper, including thanks to reviewers who gave useful comments, to colleagues who contributed to the ideas, and to funding agencies and corporate sponsors that provided financial support.}

\acknowledgments{}
This work is supported by Sony, NSF Institute for AI and Fundamental Interactions, DARPA Machine Common Sense, NSF grant 2214177, AFOSR grant FA9550-22-1-0249, ONR grant N00014-22-1-2740, MIT-IBM Watson Lab,  MIT Quest for Intelligence. Anthony Simeonov and Yilun Du are supported in part by NSF Graduate Research Fellowships. We thank members of the Improbable AI Lab and the Learning and Intelligent Systems Lab for the helpful discussions and feedback. %on the paper.

\subsection*{Author Contributions}

\textbf{Anthony Simeonov} developed the idea of minimizing descriptor variance for aligning multiple demonstrations, set up the simulation and real robot experiments, played a primary role in paper writing, and led the project.

\textbf{Yilun Du} came up with and implemented the energy-based modeling framework for relative pose inference, helped develop the overall framework of using NDFs for relational rearrangement tasks, ran simulated experiments, helped with writing the paper, and co-led the project. 

\textbf{Yen-Chen Lin} participated in research discussions about different ways to approach 6-DoF pick-and-place/rearrangement tasks, helped suggest improvements to the NDF training and optimization procedure, and helped with editing the paper.

\textbf{Alberto Rodriguez} helped with early brainstorming on how multiple NDF models could be used for multi-object rearrangement tasks and gave feedback on the tasks and real robot results.

\textbf{Leslie Kaelbling} helped develop the idea of chaining multiple pairwise relations together to perform multi-step tasks, provided suggestions on interesting rearrangement tasks to solve, and helped write and edit the paper.

\textbf{Tom\'as Lozano-Per\'ez} also helped suggest the application to multi-step tasks via sequencing relations, reinforced the investigation of representations grounded in local interactions between object parts, and provided valuable feedback on the paper.

\textbf{Pulkit Agrawal} was involved in early technical discussions about how to use multiple NDF models for rearrangement tasks, helped clarify key technical insights regarding query point labeling in the demonstrations, advised the overall project, and helped with paper writing and editing.

%===============================================================================

% \bibliographystyle{abbrv}
% {\small
\bibliography{include/example.bib}
% }

\pagebreak

\appendix
\renewcommand{\thesection}{A\arabic{section}}
\renewcommand{\thefigure}{A\arabic{figure}}
\setcounter{section}{0}
\setcounter{figure}{0}

\renewcommand{\hl}[1]{{\color{black}#1}}

% \pagebreak
% \section*{Appendix}
% \def\myformat#1{\centering#1}
\def\centsect#1{\Large\centering#1}
\section*{\centsect{SE(3)-Equivariant Relational Rearrangement with Neural Descriptor Fields: Supplementary Material}}

% \label{sec: appendix}
%%%%%%%%%%%%%%%%%%%%%%%%%%%%%%%%%%%%%%%%%%%%%%%%%%%%%%%%%%%%%%%%

In Section \ref{sect:ndf-training}, we present details on data generation, model architecture, and training for NDFs. In Section \ref{sect:ndf-optimization} we detail the optimization method used to recover the pose of a local coordinate frame by minimizing descriptor distances (as in Equations (4), (5), and (6)). Section \ref{sect:ebm-details} describes the procedure for training the energy-based models used in relative transformation refinement. In Section \ref{sect:experimental-setup}, we describe more details about our experimental setup, Section \ref{sect:eval-details} discusses more details on the evaluation tasks and robot execution pipelines, and Section \ref{sect:alternate_min} describes our alternating minimization method for aligning descriptors across a set of demonstrations. \hl{In section \ref{sect:collision-avoidance} we present an additional set of qualitative results showing how R-NDF can be used to handle collision avoidance and additional problem constraints and more complex rearrangement tasks, and in Section \ref{sect:multi-object-rearrangement} we discuss applying R-NDF to relational rearrangement with more than two objects. Section \ref{sect:partial-pointclouds} shows an example of the framework operating with partial point clouds and Section \ref{sect:misc-visualizations} contains additional visualizations of the tasks and objects used in the evaluation. Finally, in Section \ref{sect:impl-details-limits}, we provide more thorough implementation details and an expansion on the limitations of the proposed approach.}

\section{NDF Training}
\label{sect:ndf-training}

In this section, we present details on the data used for training NDFs, the neural network architectures we used in the NDF implementation, and model training.

\subsection{Training Data Generation}
\begin{figure*}[t]
\includegraphics[width=\linewidth]{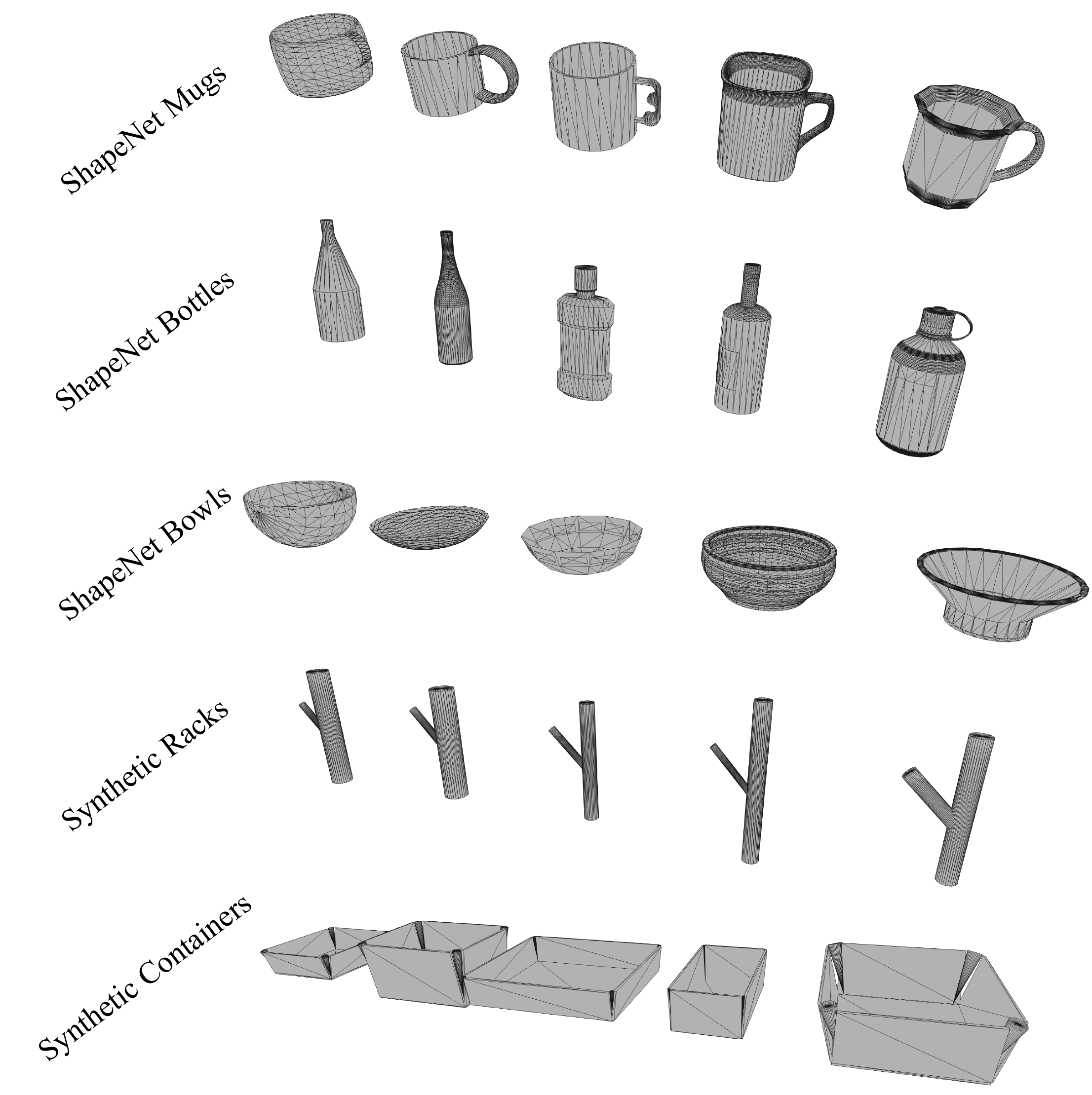}
\caption{\small  Example 3D models used to train NDFs and deploy NDFs on our rearrangement tasks. Mugs, bottles, and bowls are from ShapeNet~\cite{chang2015shapenet} while we procedurally generated our own synthetic racks and box-shaped containers.}
\label{fig:example-3d-models}
\end{figure*}

\myparagraph{3D shape data for training NDFs}
NDFs are trained to perform category-level 3D reconstruction from point cloud inputs. 
We supervise this training using ground truth 3D shape data obtained from synthetic 3D object models. 
The three tasks we consider include objects from five categories: mugs, bowls, bottles, racks, and containers.
Our NDF training thus begins with obtaining a dataset of varying 3D models for a diverse set of object instances from each of these categories.
We use ShapeNet~\cite{chang2015shapenet} for the mugs, bowls, and bottles, and procedurally generate our own dataset of \texttt{.obj} files for the racks and containers. See Figure~\ref{fig:example-3d-models} for representative samples of the 3D models from each category.

\myparagraph{NDFs based on regressing occupancy vs. signed-distance}
Given an object dataset of 3D models, we generate a dataset of inputs and outputs for training the neural networks used in NDFs.
While in~\cite{simeonov2022neural} the underlying NDF decoder is trained to perform 3D reconstruction by representing an occupancy field~\cite{mescheder2019occupancy}, i.e., as an Occupancy Network (ONet) that predicts whether a point is inside or outside of a shape, we find performance improves by instead training the model to regress a signed-distance field (SDF) ~\cite{park2019deepsdf}, i.e., as a DeepSDF that predicts the minimum distance from a point to the boundary of a shape and assigns negative/positive values for points inside/outside of the shape.
Details on our pipeline for generating the data used to train the SDF decoder can be found in the next subsection.
The following paragraphs discuss two reasons we hypothesize for the performance gap between occupancy field and signed-distance field training.

First, a signed-distance field (whose zero level set represents the boundary of a 3D shape) contains information about the underlying object geometry even at query points that are \emph{far away} from the surface of the object, whereas the underlying occupancy field is flat everywhere except exactly at the crossing between the inside and outside of the shape.
This feature of an SDF increases the likelihood of \emph{unique} descriptors at different coordinates (e.g., $\pointndf(\coord_{i}|\pointcloud)$ and $\pointndf(\coord_{j}|\pointcloud)$).
These factors appear to shape the optimization landscape in Equations (4), (5), and (6) to enable smoother and more consistent convergence, along with less sensitivity to poor initialization.

Second, in addition to an (empirically observed) improvement in global optimization convergence, we also observe the SDF-based decoder leads to the optimization performing much better \emph{near} the surface of the shape.
The intuition is that an occupancy field has a sharp discontinuity at the boundary of the shape.
When optimizing a query point set pose using Equation (4)-(6), the gradients become steep when the optimization reaches the region near the input point cloud. 
Some of the points in $\probes$ end up inside the shape and get stuck.
In contrast, although it may still have high-frequency fluctuations near the shape boundary, an SDF varies much less rapidly near the surface, and we observe a corresponding reduction in local minima when running the NDF optimization using the SDF model.

\myparagraph{Data generation}
Based on the empirical observations discussed above, we convert each 3D model dataset into a corresponding dataset of input/output pairs for signed-distance function regression.
For each shape, we normalize the object to a unit bounding box and use the 3D model to generate the distance from $M$ query points to the boundary of the shape, where points inside and outside the shape are labeled with negative and positive sign, respectively.
We use an open-source tool\footnote{\url{https://github.com/marian42/shapegan} and \url{https://github.com/marian42/mesh_to_sdf}} for generating the ground truth signed distance. We computed signed-distance values for $M=200,000$ query points per shape, where the query points are sampled inside a unit sphere and biased toward being near the surface of the shape (using about $M/2$ points near the shape boundary). 

We also need a point cloud of each shape to provide as input during training.
To generate these point clouds, we initialize the objects on a table in PyBullet~\cite{coumans2016pybullet} in random positions and orientations, and render depth images with the object segmented from the background using multiple simulated cameras. 
These depth maps are converted to 3D point clouds and fused into the world coordinate frame using known camera poses.
To obtain a diverse set of point clouds, we randomize the number of cameras (1-4), camera viewing angles, distances between the cameras and objects, object scales, and object poses.
Rendering point clouds in this way allows the model to see some of the occlusion patterns that occur when the objects are in different orientations and cannot be viewed from below the table.

\subsection{Architecture}

\myparagraph{Point cloud encoder} We follow the encoder/decoder architecture proposed in Vector Neurons~\cite{deng2021vector} for rotation equivariant occupancy networks~\cite{mescheder2019occupancy}, and replace the output occupancy probability prediction with a signed distance regression.

The encoder $\encoder$ follows from the SO(3)-equivariant PointNet~\cite{qi2017pointnet} model proposed in~\cite{deng2021vector}. 
$\encoder$ takes $\pointcloud \in \mathbb{R}^{3 \times N}$ as input and outputs a \emph{matrix} latent representation $\mathbf{Z} \in \mathbb{R}^{3 \times C}$ (i.e., to which applying a rotation $\mathbf{R} \in$~\sothree{} is a meaningful operation). 
The composition of layer operations in $\encoder$ is rotation equivariant (see~\cite{deng2021vector} for details on how this property is achieved). 
As a result, given a point cloud $\pointcloud$ and rotation $\mathbf{R}$, the following relationship holds by construction: %the encoder $\encoder$ has the property 
\begin{gather}
\encoder(\mathbf{R} \pointcloud) = \mathbf{R}\encoder(\pointcloud) = \mathbf{R} \mathbf{Z}.
\end{gather}

\myparagraph{Implicit function decoder} The decoder $\Phi$ consists of an MLP with residual connections that also contains Vector Neuron layers:
\begin{gather}
    \Phi(\coord, \encoder(\pointcloud)): \mathbb{R}^3 \times \mathbb{R}^{3 \times C} \to \mathbb{R}
    \label{eq:occupancy}
\end{gather}
The decoder takes in a combination of features, computed using the point cloud embedding $\mathbf{Z}$ and a 3D query point $\coord$, that is designed to make the output prediction rotation \emph{invariant}, i.e., if $\mathbf{Z}$ and $\coord$ have been rotated \emph{together}, the distance prediction should not change.
Specifically, following~\cite{deng2021vector}, the decoder predicts the signed distance $s \in \mathbb{R}$ from 3D coordinate $\coord$ to the shape as: 
\begin{gather}
    s(\coord, \mathbf{Z}) = \text{ResNet}(\big[ \langle \coord,  \mathbf{Z} \rangle , ||\coord||^{2}, \text{VN-In}(\mathbf{Z}) \big] ) 
\end{gather}
where $\text{VN-In}(\mathbf{Z})$ is rotation \emph{invariant}, obtained as $\text{VN-In}(\mathbf{Z}) := \mathbf{V}^{T}\mathbf{Z}$, where $\mathbf{V}$ is the output of another rotation equivariant function of $\mathbf{Z}$\footnote{Following the explanation from~\cite{deng2021vector}, this is because the product of one rotation equivariant feature $\mathbf{Z} \in \mathbb{R}^{3 \times C}$ with the transpose of another $\mathbf{V} \in \mathbb{R}^{3 \times C'}$ is rotation invariant: $(\mathbf{R}\mathbf{V})^{T} (\mathbf{R}\mathbf{Z}) = \mathbf{V}^{T}\mathbf{R}^{T} \mathbf{R}\mathbf{Z} = \mathbf{V}^{T}\mathbf{Z}$. See Sec. 3.5 of~\cite{deng2021vector} for more details}. 

\myparagraph{Building descriptors from MLP activations}
Following prior work~\cite{simeonov2022neural}, we define an NDF as a function mapping an input coordinate and point cloud to the \emph{vector of concatenated activations} of the decoder $\Phi$:
\begin{align}
    \pointndf(\coord | \pointcloud) = \bigoplus_{i=1}^L \Phi^i(\coord, \encoder(\pointcloud))
    \label{eq:pointndf}
\end{align}
where $L$ denotes the total number of layers in $\Phi$, $\Phi^{i}$ denotes the output activation of the $i$th layer, and $\bigoplus$ denotes the concatenation operator.

\subsection{Training Details}
Here we discuss details on training the NDF models using the data and model architecture described in the sections above.
Training samples consist of inputs (point cloud $\pointcloud_{train} \in \mathbb{R}^{N_{train} \times 3}$, with $N_{train}=1000$) and a set of query points $\probes_{train} \in \mathbb{R}^{N_{q, train} \times 3}$, with $N_{q, train} = 1500$) and ground truth outputs (distance between each point in $\probes_{train}$ and the underlying shape of the object represented by $\pointcloud_{train}$).
Based on the corresponding object scale and pose used to generate the respective point cloud in simulation, we compute ground truth distances by transforming and scaling the distances relative to the normalized object shapes. 
We train $\encoder$ and $\Phi$ end-to-end to minimize the mean-squared error between the predicted and ground-truth distance values across the query points.

We generated a dataset of approximately 100,000 samples for each category and trained one NDF model per category using each respective dataset.
We trained the models for 150 thousand iterations on a single NVIDIA 3090 GPU with a batch size of 16 and a learning rate of 1e-4, which takes about half a day. We train the models using the Adam~\cite{kingma2014adam} optimizer.

\section{NDF Coordinate Frame Localization via Descriptor Distance Minimization}
\label{sect:ndf-optimization}
This section describes how the optimization in Equation (4) is performed to recover a pose, relative to an unseen shape, that matches a demonstration pose which has been encoded into a target pose descriptor.

The inputs to the problem are the target pose descriptor $\hat{\posedescriptor}$, NDF $\pointndf$, point cloud $\pointcloud$, and query points $\probes$ (in their canonical pose).
We randomly initialize an \sethree{} pose $\pose^{0}$ as an axis-angle 3D rotation $\rotaa^{0} \in \mathbb{R}^{3}$ and a 3D translation $\mathbf{t}^{0} \in \mathbb{R}^{3}$. 
We then perform $N_{iter}$ iterations of gradient descent to update this pose.

On iteration $j$, we begin by constructing a pose $\pose^{j} \in$ \sethree{} by converting $\rotaa^{j}$ into rotation matrix $\mathbf{R}^{j} \in$ \sothree{} using the \sothree{} exponential map~\cite{sola2018micro} and combining with the translation $\mathbf{t}^{j}$.
We then obtain pose descriptor $\posedescriptor^{j}$ by transforming the query points $\probes$ by $\pose^{j}$ and applying Equation (3) using $\pointndf$.
Finally, we compute the loss $\mathcal{L}$ using the L1 distance between $\hat{\posedescriptor}$ and $\posedescriptor^{j}$ and backpropagate gradients of this loss to update the rotation and translation:
\begin{gather}
    \rotaa^{j+1} \leftarrow \rotaa^{j} - \lambda \nabla_{\rotaa} \mathcal{L} \\
    \mathbf{t}^{j+1} \leftarrow \mathbf{t}^{j} - \lambda \nabla_{\mathbf{t}} \mathcal{L}    
\end{gather}
We use the Adam~\cite{kingma2014adam} optimizer with a learning rate $\lambda$ of 1e-2 to run this procedure. 
The optimization is run for a fixed number $N_{iter} = 650$ iterations to optimize $\pose$. We empirically observed this to be enough iterations to allow the solution to converge.
Furthermore, since the loss landscape for this problem is non-convex, the solution is somewhat sensitive to the initialization.
To help obtain some diversity in the solutions, we run the optimization multiple times in parallel from different initial values for the pose.
We used a batch size of 10, which uses about 10GB of GPU memory when using $N_q=500$ query points and a point cloud downsampled to $N=1500$ points.

\section{EBM Training}
\label{sect:ebm-details}

In this section, we present the training details of utilizing an EBM to refine predictions of relative transformation between objects. 

\myparagraph{Training Data} To train EBMs to capture each relation, we utilize the 10 demonstrations provided for each task. \hl{To prevent overfitting, and to construct more diverse data to train EBM models, we heavily data augment point clouds in each demonstration. In particular, we skew point clouds, apply per-point Gaussian noise, and simulate different occlusion patterns on the demonstration point clouds}.

\myparagraph{Model Architecture} We utilize a three-layer MLP (described in Table \ref{tbl:ebm}) to parameterize an EBM operating over the concatenation of descriptors at each point. We utilize a swish activation in the EBM to enable fully continuous gradients with respect to inputs.

\myparagraph{Training Details} When training EBMs, we corrupt $\demopose_{B}\demopointcloud_B$ by a transformation corresponding translation sampled from $[-0.05, 0.05]$ along each dimension and rotation perturbation of 15 degrees along yaw, pitch and roll. We utilize a gradient descent step size of $10$ for translation and $20$ for rotation during optimization in training, and run 8 steps of optimization. Optimized rotations are represented as Euler angles, as the perturbations of individual rotation components are small. Each EBM is evaluated pointwise across 1000 points (with descriptors with respect to each object concatenated).  EBMs are trained with a batch size of 16.

\myparagraph{Computational Resources} To train each model, we utilize a single Volta 32GB machine for 6 hours and train models for 12000 iterations.

\section{Experimental Setup}
\label{sect:experimental-setup}
This section describes the details of our experimental setup in simulation and the real world.

\subsection{Simulated Experimental Setup}
We use PyBullet~\cite{coumans2016pybullet} and the AIRobot~\cite{airobot2019} library to setup the tasks in the simulation, provide demonstrations, and perform quantitative evaluation experiments.
The simulation environment contains a Franka Panda arm with a Robotiq 2F140 gripper attached, a table, and a set of simulated RGB-D cameras.
We obtain segmentation masks using the built-in segmentation abilities of the simulated cameras to separate object point clouds from the overall scene.

\subsection{Real World Experimental Setup}

Our real-world environment also contains a Franka Robot arm with a Robotiq 2F140 parallel jaw gripper.
We also use four Realsense D415 RGB-D cameras, with extrinsics calibrated relative to the robot's base frame. 
We use a combination of point cloud cropping and Euclidean clustering to segment object point clouds from the scene, identify their class identities, and filter out noise.
Specifically, we crop the overall scene to the known region above the table, and then crop $\object_{A}$ and $\object_{B}$ based on which side of the table each object is on (assumed to be known for this experiment, just for the purposes of obtaining the segmentation). 
Finally, we run DBSCAN~\cite{ester1996density} clustering to remove outliers and noise to obtain the final point clouds $\pointcloud_{A}$ and $\pointcloud_{B}$.
To demonstrate \ourshort{} on objects in diverse initial orientations, we present some of the objects on a 3D-printed stand with angled support surfaces.
When using the stand, we remove it from the point cloud by estimating its pose using 3D registration and the known CAD model.

\section{Evaluation Details}
\label{sect:eval-details}
This section presents further details on the three tasks we used in our experiments and notes on the automatic detection methods used to obtain success rates over multiple simulated trials.

\subsection{Tasks and Evaluation Criteria }
\myparagraph{Task Descriptions}
We consider three relational rearrangement tasks for evaluation: (1) Hanging a mug on the hook of a rack, (2) Stacking a bowl upright on top of a mug, and (3) Placing a bottle upright inside of a box-shaped container.
For ``Hanging a mug on a rack'', we ensure the mug is oriented consistently relative to the single peg of the rack. This is achieved by providing demonstrations in which the handle always points left relative to a front-facing peg, and the opening of the mug is always points toward the top of the rack. 
Similarly for stacking bowls on mugs and putting bottles in containers, we provide demonstrations in which the ``stacked'' object is always upright, though in these tasks, the orientation about the radial axis of the bowls/bottles doesn't affect the relation result and is thus ignored. 
All objects are presented in an initial ``upright'' pose on the table for the demonstrations. 

\myparagraph{Evaluation Metrics and Success Criteria}
To quantify performance, we report the success rate over 100 trials, where we use the ground truth simulator state to compute success.
For a trial to be successful, objects $\object_{A}$ and $\object_{B}$ must be in contact, $\object_{B}$ must have the correct orientation relative to $\object_{A}$, and $\object_{A}$ and $\object_{B}$ must not interpenetrate.

\subsection{Providing Demonstrations}
Here we re-describe the procedure for obtaining task demonstrations using a robot arm, a set of depth cameras, and a parallel jaw gripper, as outlined in Section 5.

When collecting a demonstration, initial object point clouds $\demopointcloud_{A}$ and  $\demopointcloud_{B}$ of objects $\demoobject_{A}$ and $\demoobject_{B}$ are obtained by fusing a set of back projected depth images. 
The demonstrator moves the gripper to a pose $\demopose_{\text{grasp}}$, grasps $\demoobject_{B}$, and finally moves the gripper to a pose $\demopose_{\text{place}}$ that satisfies the desired relation between $\demoobject_{A}$ and $\demoobject_{B}$.
$\demopose_{B}$ is obtained as $\demopose_{\text{place}}\demopose_{\text{grasp}}^{-1}$. % %
In \emph{one} of the demonstrations, a 3D keypoint $\coord_{AB}$ is labeled near the parts of the objects that interact with each other by moving the gripper to this region and recording its position. 

\hl{
\myparagraph{Tuning the scale of the query point set $\mathcal{X}_{A}$}
As mentioned in Section 4.1, $\probes_{A}$ is obtained by sampling from a zero-mean Gaussian, with a variance that must be chosen by the user. This variance will impact the scale of the query point set, and can be thought of as a hyperparameter. The original NDF paper discusses the implications of different query point sizes (see Table III in \cite{simeonov2022neural}). We therefore did not repeat the ablations performed in \cite{simeonov2022neural} showing that the scale of the query point cloud must be tuned based on the rough scale of the shapes in the object set. We instead tuned the variance parameter and settled on values that led to good task performance. For real-world objects, the value we used typically falls between 0.015 and 0.025. The heuristic used in \cite{simeonov2022neural} to simplify this tuning procedure was to use a bounding box around the whole shape $\object_{A}$.
}

\subsection{Baselines}
\myparagraph{Pose Regression} The Pose Regression baseline method consists of an MLP trained to predict $\pose_{B}$ using the concatenated embeddings of $\pointcloud_{A}$ and $\pointcloud_{B}$ obtained from a pretrained VNN encoder (the one used for NDFs). We supervise transform prediction $\hat{\pose}_B$ using the Chamfer distance between a transformed point cloud $\hat{\pose}_B \pointcloud_B$ and a ground truth point cloud $\pose_B \pointcloud_B$. Such a loss captures the symmetry in $\pointcloud_B$. A transformation is represented using vector of six dimensions, where the first three dimensions correspond to translation and the subsequent dimensions correspond to an axis-angle parameterization of rotations  The architecture for pose regression is provided in Table \ref{tbl:baseline_pose_regress}. 

\myparagraph{Patch Match} The Patch match baseline uses 3D registration to align the test-time point clouds $\pointcloud_{A}$ and $\pointcloud_{B}$ to the point clouds of the corresponding shapes used in one of the demonstrations, and then using the demonstrated pose $\hat{\pose_{B}}$ together with these registration results to compute the resulting pose $\pose_{B}$. 
Formally, for demonstration $\mathcal{D}_{i}$, registering source $\pointcloud_{A}$ to target $\demopointcloud_{A}$ produces \sethree~transformation $\pose_{A, \text{reg}}$, and similarly for $\pointcloud_{B}$ and $\demopointcloud_{B}$ to obtain $\pose_{B, \text{reg}}$. %
$\pose_{B}$ is then obtained as $\pose_{A, \text{reg}}^{-1}\demopose_{B}$. 

\subsection{Automatic success detection and common failure modes} This section discusses the methods we used for checking each success criteria in the simulator, along with some of the common failure modes for each method. 

\myparagraph{Correct Orienttion} We compare the angle between the radial axis of the bowls/bottles and the positve z-axis in the world to check if the final orientation is correct for the ``bowl on mug'' and ``bottle in container'' tasks. We count the objects as ending in a valid ``upright'' orientation if this angle difference is below 15 degrees once the physics has been turned on and the object has settled.
This is important because a common failure mode for the ``bottle in container'' task is to localize the top of the bottle instead of the bottom and try to place it upside down on the container. This occurs for both our method and the baselines. For ``bowl on mug'', this failure mode is much less apparent for our method but still occurs quite often with the baselines.

We did not explicitly check if the orientation is correct for the ``mug on rack'' task, as this would require comparing the angle between the axis pointing along the cylindrical body of the mug and the axis pointing along the rack's peg to ensure the mug opening points in the right direction relative to the rack. Since the pegs on the racks are all slightly different, obtaining this ground truth peg-axis angle was too cumbersome to manually set up. We instead relied on the ``$\object_{A}$ and $\object_{B}$ in contact'' criteria to imply the correct relative configuration between the mug and the rack. This is an effective method because for many incorrect relative orientations, the mug misses the rack and falls when the physics are turned on, leading to a final configuration where the objects do not touch. However, there may still be solutions found that satisfy the ``hanging'' criteria by passing this check but not being in the intended orientation. We thus allow any final configurations satisfying ``hanging'' (implied by the mug and rack ending up ``in contact'' and ``not interpenetrating''), where the mug opening points down instead of up, to be counted as a success. We observe this happens much more frequently for the ``Patch Match'' baseline than \ourshort{}, as ``Patch Match'' struggles in disambiguating between an upright and an upside-down mug during 3D registration. 

\myparagraph{In-Contact vs. Not-Interpenetrating} Since we reset $\object_{A}$ and $\object_{B}$ in the simulator to their final configuration after predicting the relative transformation, the objects can end up in physically-/geometrically-infeasible poses that intersect, and we don't want to count these configurations as successful for any of the tasks. As described above, we use the ``in-contact'' criterion to implicitly determine if the ``mug on rack'' relation is satisfied based on the objects' relative orientation. Even for ``bottle in container'' and ``bowl on mug'', where we explicitly check to ensure $\object_{B}$ ends in an upright orientation, we might obtain a prediction that places $\object_{B}$ too low relative to $\object_{A}$ and causes an infeasible intersection. Therefore, we can obtain many false positives if we don't take care to ensure objects that interpenetrate are not counted as being \emph{successfully} ``in-contact''. However, automatically disambiguating the ``in contact'' criteria with the ``not interpenetrating'' criteria, both of which are required for success, turns out to be slightly nontrivial. Here we discuss the method we used to check these criteria in a disentangled fashion.

We first check that $\object_{A}$ and $\object_{B}$ are in contact after transforming $\object_{B}$ by $\pose_{B}$ (``in contact'') and allowing the physics simulation to proceed for 2 seconds,.
We then ensure the objects are not only in contact because they are interpenetrating by checking whether or not $\object_{B}$ can be easily \emph{removed} from its final configuration under dynamic physical effects.
To achieve this, we transform $\object_{A}$ and $\object_{B}$ to maintain their relative configuration, but make $\object_{A}$ turn upside down in the world frame. 
For each of our tasks, if the objects are not in interpenetration, $\object_{B}$ should fall away from $\object_{A}$, whereas we observe that they regularly get stuck together in a physically implausible way if they are in interpenetration.
We thus check whether or not the objects are still in contact after turning them upside down and waiting while the physics simulation proceeds, and label the pair as ``not interpenetrating'' if they are not in contact after this delay.
The most common failure mode for \ourshort{}s on the ``bottle in container'' and ``bowl on mug'' tasks is to predict transformations that nearly satisfy the relation but cause the objects to interpenetrate (e.g., the bottle/bowl is too low, and thus intersects with the container/mug).

\subsection{Task Execution}
This section describes additional details about the pipelines used for executing the inferred relations in simulation and the real world. 

\myparagraph{Simulated Execution Pipeline} 
The evaluation pipeline mirrors the demonstration setup.
Objects from the 3D model dataset for the respective categories are loaded into the scene with randomly sampled position and orientation.
In the ``upright'' pose case, a known upright orientation is obtained, and then adjusted with a random top-down yaw angle.
In the ``arbitrary'' pose case, we sample a rotation matrix uniformly from \sothree, load the object with this orientation, and constrain the object in the world frame to be fixed in this orientation. 
We do not allow it to fall on the table under gravity, as this would bias the distribution of orientations covered to be those that are stable on a horizontal surface, whereas we want to evaluate the ability of each method to generalize over all of \sothree.
In both cases, we randomly sample a position on/above the table that are in view for the simulated cameras.
We also load the target pose descriptor $\hat{\posedescriptor}$, obtained by following the procedure described in Section 4, for use in inference.

After loading objects $\object_{A}$ and $\object_{B}$ and the target pose descriptor $\hat{\posedescriptor}$, we obtain segmented point clouds $\pointcloud_{A}$ and $\pointcloud_{B}$ and apply Equations (5), (6), and (8) to obtain a transformation $\pose_{B}$. 
The output transformation is applied to $\object_{B}$ by transforming its initial pose $\pose_{B, \text{start}}$ by $\pose_{B}$ and resetting the state of the object in the simulation to the resulting pose $\pose_{B, \text{final}} = \pose_{B} \pose_{B, \text{start}}$. 
Task success is then checked based on the criteria described in the section above.

\myparagraph{Real World Execution Pipeline} 
Here, we repeat the description of how we execute the inferred transformation using a robot arm with additional details.

At test time, we are given point clouds $\pointcloud_{A}$ and $\pointcloud_{B}$ of new objects $\object_{A}$ and $\object_{B}$, each with potentially new shapes and poses, and Equations (5), (6), and (8) are applied in sequence to obtain $\pose_{B}$. 
We first obtain an initial grasp pose $\pose_{\text{grasp}}$.
Our implementation uses NDF to generate these grasp poses, following the pipeline described in~\cite{simeonov2022neural} and Section 3 (where $\object_{A}$ is the robot's gripper), but any generic grasp generation pipeline could be used instead.
We then obtain a placing pose $\pose_{\text{place}} = \pose_{B}\pose_{\text{grasp}}$, and plan a collision-free path between the grasp and place pose using MoveIt!\footnote{\url{https://moveit.ros.org/}}.
The path is executed by following the joint trajectory in position control mode and opening/closing the fingers at the correct respective steps.
The whole pipeline can be run multiple times in case the planner returns infeasibility, as the inference methods for both grasp and placement generation can produce somewhat different solutions depending on how the NDF optimization is initialized.

\myparagraph{Pipeline for Executing Multiple Relations in Sequence}
Figure~1 shows a multi-step rearrangement application of \ourshort{}s for the ``bowl on mug'' task, where a ``mug upright on the table'' relation is executed before the ``bowl upright on the mug'' relation. 
This section describes the setup for chaining these relations and executing them in sequence (see also Subsection~\ref{sect:multi-object-rearrangement} below for further discussion).

To specify the ``mug upright on table'' component of the task, we follow~\cite{simeonov2022neural} and the steps described in Section 3 on ``NDFs for Encoding Single Unknown Object Relations'', where the table is the \emph{known} object $\object_{A}$.
Using this prior knowledge, we initialize a set of query points near a known placing region on the table, and use these points to obtain the target pose descriptor from a set of demonstrations (i.e., demos of placing mugs upright on the table near the query point set location).
The ``bowl upright on mug'' part of the task is then encoded using the method described in Section 4.

During execution, both inference steps are run using the \emph{initial} point clouds $\pointcloud_{A}$ and $\pointcloud_{B}$, i.e., we don't re-observe the mug after executing the upright placement on the table.
Thus, we find $\pose_{B}^{1}$ which transforms the mug relative to the table, and $\pose_{B}^{2}$, which transforms the bowl relative to the mug in its \emph{initial} configuration.
Finally, we execute relative transformations $\pose_{B, \text{exec}}^{1} = \pose_{B}^{1}$ to the mug and $\pose_{B, \text{exec}}^{2} = \pose_{B}^{1} \pose_{B}^{2}$ to the bowl, using the pick-and-place operation described in the section above.

\section{Alternating Minimization to Obtain an Average Pose Descriptor From Multiple Unaligned Demonstrations}
\label{sect:alternate_min}
This section describes details and further intuition behind our method for aligning descriptors across a set of demonstrations, as proposed in Section 4.

Assuming a specified interaction point $\coord_{AB}$ in demonstration $\mathcal{D}_{i}$, we first construct $\demopose_{\probes_{A}, i}^{0}$.
We then transform the canonical query points $\probes_{A}$ by $\demopose_{\probes_{A}, i}^{0}$ to the region near the task-relevant features on the object and obtain $\hat{\posedescriptor}^{0}_{ref} = \hat{\posedescriptor}^{0}_{i}$ with Equation (3).
Equation (5) is then used with $\hat{\posedescriptor}^{0}_{ref}$ to solve for a corresponding transformation $\demopose_{\probes_{A}, j}$ and a resulting pose descriptor $\hat{\posedescriptor}^{0}_{j} = F_{A}(\demopose_{\probes_{A}, j} | \demopointcloud_{A, j})$ for the remaining demonstrations $\{\mathcal{D}_{j}\}_{j=1, j \neq i}^{K}$. 
After running this for each demonstration, we compute an average pose descriptor $\hat{\posedescriptor}^{1}_{ref} = \frac{1}{K} \sum_{i=1}^{K} \hat{\posedescriptor}^{0}_{j}$. 
We then apply the same procedure to each demonstration (including the demo used for providing the initial reference pose) \emph{again}, now using $\hat{\posedescriptor}^{1}_{ref}$ as the target descriptor. We repeat this process $Q$ times, where $Q$ is a hyperparameter (we used $Q=3$ throughout the experiments).

Intuitively, this procedure starts with a target descriptor obtained from a single demonstration (whichever demonstration $\mathcal{D}_{i}$ corresponds to the one where $\coord_{AB}$ was provided). 
As shown in the top row of Table 5a, some fraction of the poses obtained by matching a single demonstration correctly match a target descriptor ($\sim$ 40\% success). 
This means that \emph{some} of the individual descriptors in $\{\hat{\posedescriptor}^{k}_{j}\}_{j=1, j \neq i}^{K}$ found on the $k$th iteration correctly align with the reference, and are somewhat near each other in descriptor space. 
The mean of this set $\hat{\posedescriptor}^{k+1}_{ref}$ thus provides a new descriptor that is both (on average) more similar to each of the descriptors than was the original reference, and still similar to the original descriptor that was obtained using the manually specified keypoint. 
By resetting the target using this mean, the next alignment round uses a target that is more sensitive to the part features that are shared among \emph{some} of the demonstrations, and makes it more likely that a consistent pose will be found for \emph{more} of the demonstrations than the previous iteration.
The similarity among the resulting descriptors continues to increase accordingly.
Overall, multiple rounds of this procedure lead to poses for each respective demonstration which are consistent with each other and map to descriptors with relatively high similarity. 

By encouraging the individual descriptors in the set to converge to values that are similar to the mean among the whole set, this iterative procedure allows the final target descriptor to capture the shared parts of the demonstration objects in a consistent fashion. 

% \color{blue}
\section{Modifying the R-NDF Optimization Objective for Collision Avoidance}
\label{sect:collision-avoidance}
\begin{figure*}[t]
\includegraphics[width=\linewidth]{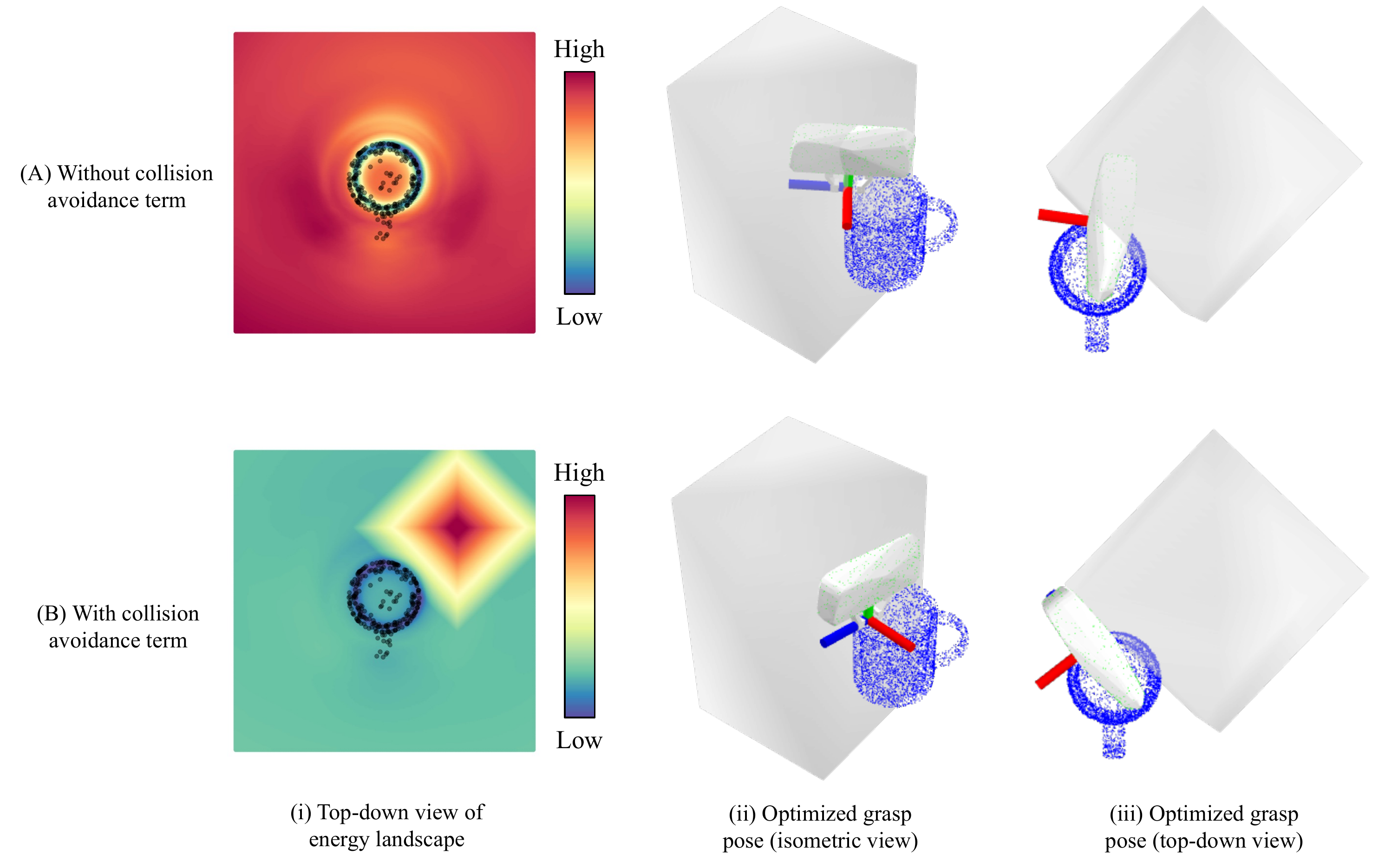}
\caption{\hl{ \small \textbf{Energy composition for collision avoidance and descriptor matching.} \textbf{(Top) NDF grasp pose inference without collision avoidance cost.} We recorded a grasp pose along the rim of a demo mug and recovered the corresponding pose on a new mug using NDF optimization. This procedure ignores collision avoidance with nearby obstacles, and thus finds a solution that would be infeasible due to interpenetration with the box-shaped obstacle next to the mug. \textbf{(Bottom) NDF grasp pose inference with collision avoidance cost.} We modify the NDF optimization with an additional cost term. This extra term penalizes query points that fall inside the obstacle. The resulting energy landscape takes both the collision avoidance and the descriptor matching into account. The optimizer finds a new solution that stays outside of the obstacle, while still achieving a grasp along the rim.} }
\label{fig:coll-avoid}
\end{figure*}
This section shows how the optimization objective used for recovering coordinate frame poses can be modified to take collision avoidance into account.

Our framework models rearrangement tasks in the form of SE(3)-transformations of rigid bodies. The method described in Section 4 does not address other constraints like collision avoidance and robot kinematics, and our real-world task executions depend on a separate motion planning module, since our approach does not produce full paths/trajectories.
However, because we perform optimization for pose localization, it is straightforward to incorporate extra cost terms that capture additional factors like collision avoidance. 

In Figure~\ref{fig:coll-avoid}, we highlight this ability with an example of grasping a mug while avoiding collisions. The top row optimizes the standard descriptor matching objective and doesn't consider nearby obstacles. In contrast, the bottom row shows the result when the optimizer minimizes a combined energy, composed of a descriptor-matching cost and a collision avoidance cost. The grasp pose found in the top row collides with the nearby obstacle, since it only tries to match the target pose descriptor. On the other hand, in the bottom row, the energy landscape for the optimization takes the nearby obstacle into account, such that a different point along the rim of the mug achieves the new optimum.
The new optimum still satisfies the desired grasp along the rim but also achieves the secondary constraint of keeping the gripper outside of the box-shaped obstacle.

This highlights the ability to incorporate additional task constraints together with the pose-matching objective, using collision avoidance as one such example. The same principle can apply to other task constraints. For instance, recent work has shown that energy-based modeling is useful for taking robot kinematics into account by setting up joint space decision variables and using a differentiable forward kinematics module~\cite{ganapathi2022implicitkinematic}. In this way, a user can add a similar additional cost term that penalizes solutions that cause self-collisions and violate joint limits. Other recent work has set up similar compositional energy optimization approaches for trajectory synthesis and motion generation~\cite{urain2021composable}. In machine learning, energy-based models have been useful for compositional generative modeling and reasoning ~\cite{du2020compositional,du2021comet,liu2022compositional,liu2021learning}.

\section{Extending Beyond Two-Object Rearrangement}
\label{sect:multi-object-rearrangement}
This section expands on how R-NDF can be used for rearrangement tasks involving more than two objects, via both sequential and compositional optimization.

Our approach models relational rearrangement via pair-wise relations, which facilitates a natural way to go beyond two-object rearrangement and handle tasks involving more objects. Here we discuss two approaches for achieving this with R-NDF. While we did not use the EBM for these examples, the same type of compositionality and sequential inference is also directly applicable to using the learned model, which also operates on pairs of object point clouds.

\subsection{Three-body rearrangement via sequential optimization}
\begin{figure*}[t]
\centering
\includegraphics[width=\linewidth]{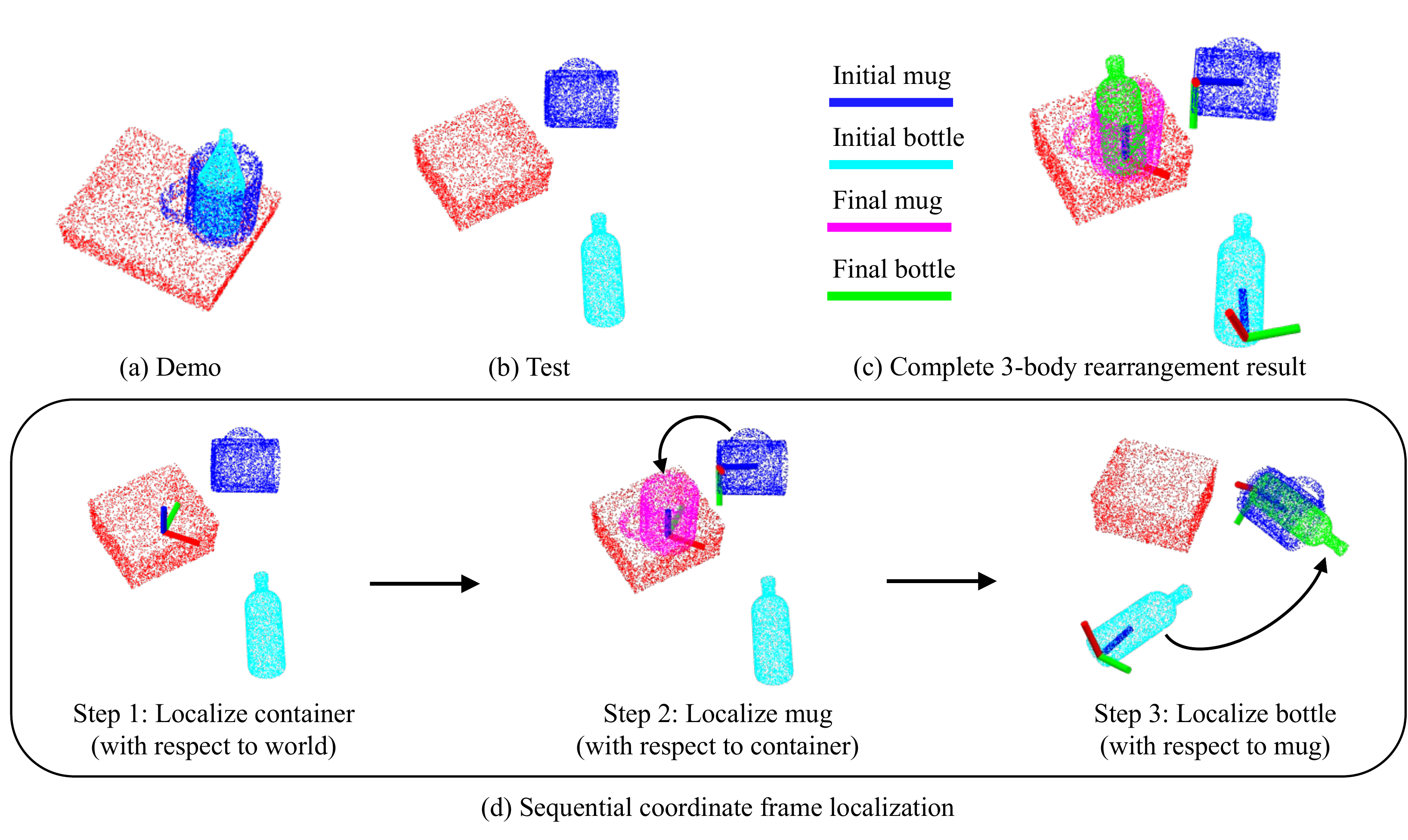}
\caption{ \hl{ \small \textbf{Three-object rearrangement by sequentially localizing multiple task-relevant coordinate frames}. \textbf{(a)} Demonstration of ``mug in container'' and ``bottle in mug''. \textbf{(b)} Initial configuration of test-time container, bottle, and mug. \textbf{(c)} Final configuration satisfying the desired set of relations among the three objects after inferring task-relevant coordinate frames on each of them. \textbf{(d)} Independent coordinate frame localization steps executed in sequence. First, the bottom of the container is found in the world frame. Then, the bottom of the sideways blue mug is found. Using the frame found in Step 1., the initial mug (dark blue) can be transformed to the final upright mug (pink). Finally, the bottom of the initially sideways bottle (teal) can be found, and transformed to the final bottle (green) inside of the mug. Composing each of these steps together provides the final configuration shown in \textbf{(c)}. } }
\label{fig:multibody-sequence}
\end{figure*}
If a set of multi-object relations forms a tree structure, each coordinate frame can be found independently and then composed in sequence to satisfy the overall task. This example is highlighted in Figure 1 (where the table can be considered a third static object). For additional clarity, Figure~\ref{fig:multibody-sequence} shows another example of this behavior. The task is to place the ``mug upright in the container \emph{and} the bottle in the mug''. At test time, the new container, mug, and bottle are all in different initial poses. First, the container is localized. Then, the mug is localized relative to the container, and finally, the bottle is localized relative to the mug. These relative transformations are composed to obtain the overall transformation applied to each object in execution.

\subsection{Three-body rearrangement via energy composition}
\begin{figure*}[t]
\centering
\includegraphics[width=\linewidth]{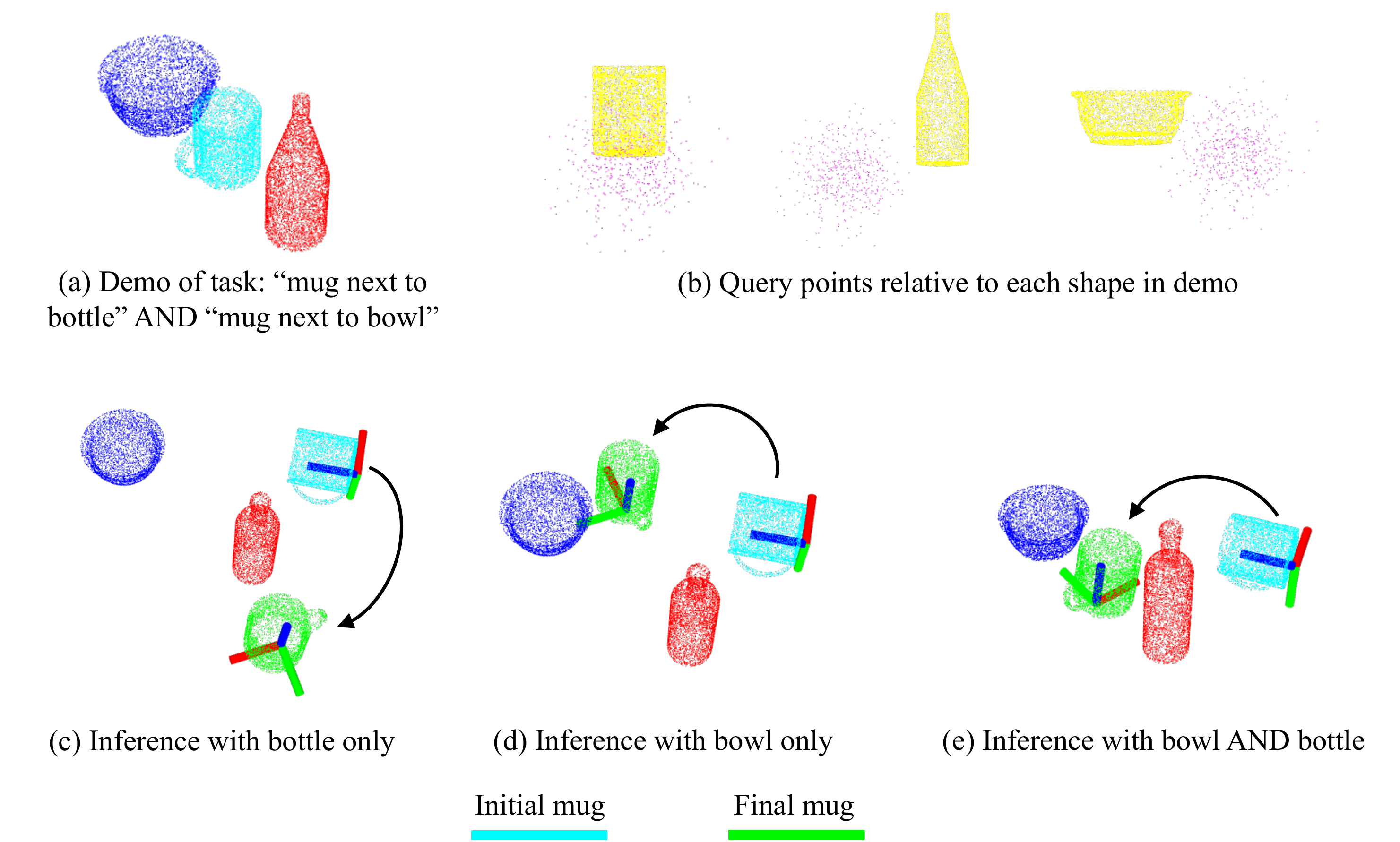}
\caption{ \hl{ \small \textbf{Three-object rearrangement by composing multiple NDF descriptor distances}. \textbf{(a)} Demonstration showing ``mug next to bottle'' \emph{and} `mug next to bowl''. \textbf{(b)} Using a single set of query points at the bottom of the demonstration mug, we obtain a set of descriptors for each shape, relative to the point cloud of each shape shown in the demonstrations. \textbf{(c)} At test-time, we first localize the world frame pose of the bottom of the mug. If we then optimize the final mug pose by matching the bottle NDF descriptors alone, a solution far away from the bowl is found. This is because the bottle has a radial symmetry which allows multiple solutions for descriptor matching. \textbf{(d)} Similarly, when finding the mug pose relative to the bowl on its own, the mug ends up at a position away from the bottle. \textbf{(e)} By optimizing the mug pose relative to both the bowl NDF and the bottle NDF \emph{simultaneously}, the resulting solution is ``next to'' both the bowl \emph{and} and the bottle.} }
\label{fig:multibody-composition}
\end{figure*}

Solving for task-relevant coordinate frame poses using optimization makes it straightforward to incorporate additional costs. We can apply this fact to the setting of three-object rearrangement, as shown in Figure~\ref{fig:multibody-composition}. In particular, instead of matching descriptors with respect to one object, we can solve for a pose that attempts to match descriptors with respect to \emph{two} objects \emph{simultaneously}. 

Here, the task is to place a mug upright both ``next to the bowl'' and ``next to the bottle'' (see Figure~\ref{fig:multibody-composition}, top left). After localizing a coordinate frame on the bottom of a mug, we run the NDF optimization of the query points to recover a coordinate frame that satisfies both of the “next to” relations simultaneously. This is achieved by computing the overall optimization loss as the sum of the two separate “bowl” and “bottle” descriptor matching losses.
Note that due to the radial symmetry of the bowl and the bottle, if we only optimize the mug pose with respect to one object, the solution can end up anywhere along a circle surrounding the respective object. However, when we optimize with respect to both objects, the solver finds a unique solution that satisfies both ``next to'' objectives, placing the mug in a location that is unambiguously between the bowl and the bottle. 

\section{Can R-NDF work with partial point clouds?}
\label{sect:partial-pointclouds}
In this section, we provide evidence that R-NDF can work with partial point clouds obtained from a single viewpoint, and discuss the difficulties of handling the problem of rearrangement with heavy occlusions in its full generality.

\begin{figure*}[t]
\includegraphics[width=\linewidth]{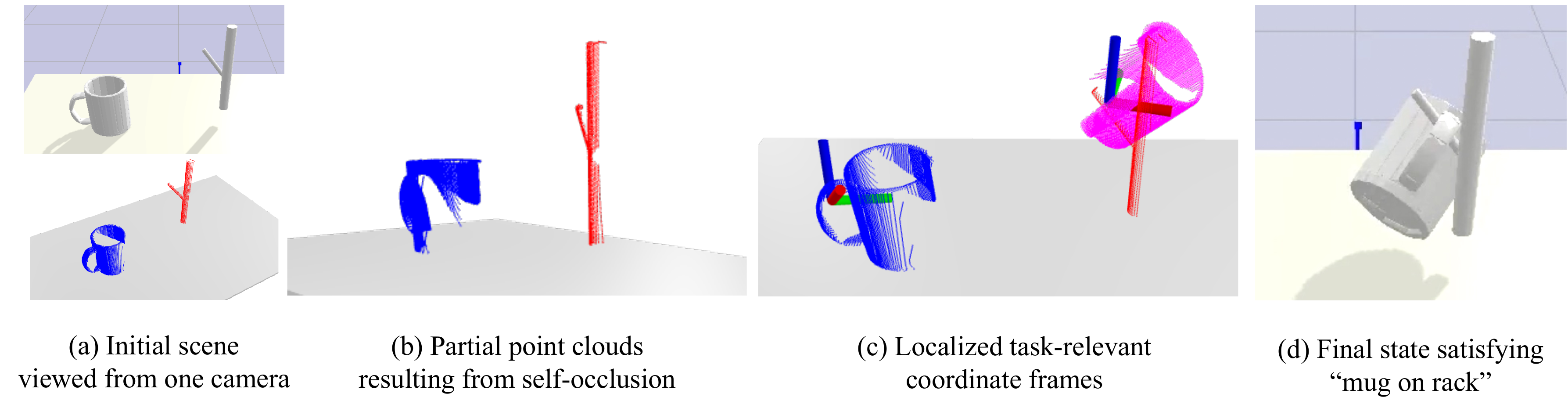}
\caption{ \hl{ \small \textbf{NDFs can work with partial point clouds.} \textbf{(a)} Initial scene for ``mug on rack'' task viewed from only a single camera. \textbf{(b)} Alternate view of the resulting point clouds for each object, showing large missing regions of each point cloud. \textbf{(c)} Task-relevant coordinate frames found on each object using the partial point clouds. This shows that, as long as the occlusions aren't too severe, NDF descriptor matching can still work for recovering corresponding coordinate frames that match a demonstration. \textbf{(d)} Final simulator state after applying the recovered relative transformation. } }
\label{fig:partial-pointcloud}
\end{figure*}

Similar to~\cite{simeonov2022neural}, we obtained point clouds from multiple cameras at different viewing angles to ensure the point cloud was relatively complete. The purpose of this design choice was to focus the effort on expanding the rearrangement task capabilities, rather than handling scenes with large occlusions. However, we have observed the model’s ability to deal with partial point clouds, so long as (1) they are included in the training distribution, and (2) the important part of the object for matching (i.e., the handle, or the peg, for the mug/rack task) is not entirely hidden from view. Figure~\ref{fig:partial-pointcloud} shows an example of successful coordinate frame localization on point clouds obtained from a single camera. 

Dealing with the partial point cloud issue in its entirety requires addressing the possibility that the task-relevant part might be completely hidden from view. This difficult scenario greatly complicates matters. Techniques related to active perception, wherein the system can search for new viewpoints, are perhaps the fundamentally correct way to deal with this problem. While this was beyond the scope of our work, note that as these complementary methods improve, our proposed method will become applicable to more general occlusion settings.

\section{Miscellaneous Visualizations}
\label{sect:misc-visualizations}
\begin{figure*}[t]
\includegraphics[width=\linewidth]{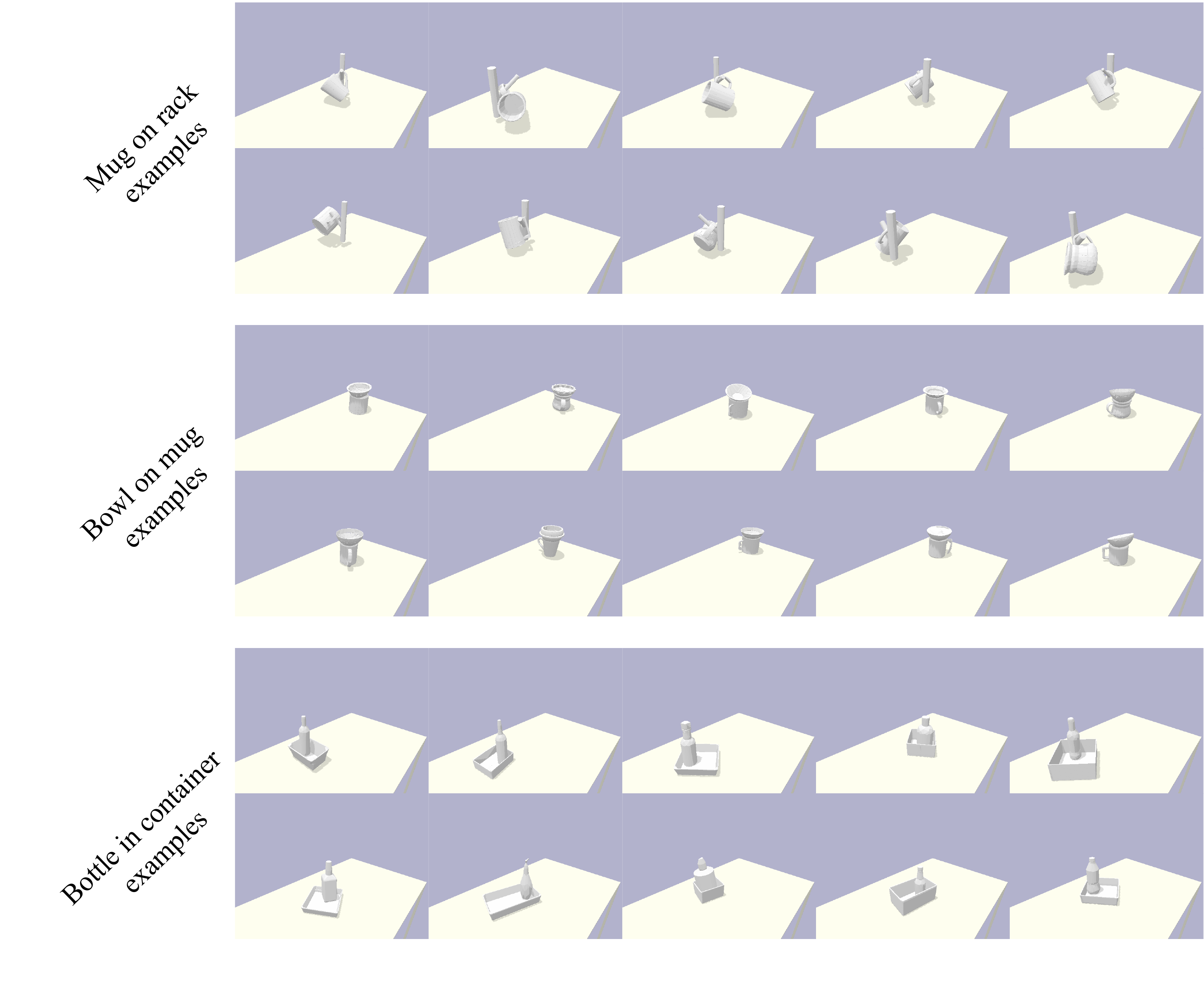}
\caption{ \hl{ \small \textbf{Additional visualizations of unseen objects for evaluation tasks.} Snapshot of the simulator final state for each evaluation task. Simulator state is reached after transforming $\object_{B}$ into its relation-satisfying configuration using R-NDF. Each image shows a successful execution of the task.} }
\label{fig:extra-qualitative}
\end{figure*}

In this section, we show additional visualizations of the tasks used in our simulation experiments. Figure~\ref{fig:extra-qualitative} shows more snapshots of the final simulator state for each of the tasks used in the quantitative evaluation. 

\section{Expanded Details on System Engineering, Implementation, and Limitations}
\label{sect:impl-details-limits}
This section provides more details on our assumptions, how the overall system is engineered, and the resulting implications and limitations. Section 8 discusses some of these considerations, and we provide a more thorough treatment here. 

\subsection{Real World Systems Engineering and Implementation}
\myparagraph{Motion planning and collision avoidance in real-world experiments}
This work models relational rearrangement in the form of relative SE(3) transformations. We focus on evaluating whether or not the inferred transformation achieves a desired relation between pairs of unseen objects. In simulation, we can directly evaluate this core capability by bypassing the physics and resetting the object state to its predicted goal configuration. However, in the real world, we must execute a feasible path to the goal with the object held by the robot's end-effector. 

Naive trajectory generation (e.g., via linear interpolation in task or joint space) regularly fails to achieve this because the robot or the grasped object collides with part of the environment. For example, hanging the mug on the rack requires a particular path to be followed in the last few inches to avoid moving the rack (e.g., by the mug colliding with the peg). 
Solving this motion planning and collision avoidance issue in its full generality would require performing full collision-free planning of the arm with the grasped object, which remains a challenging task in itself when dealing with raw point clouds. Several recent works~\cite{you2021omnihang, danielczuk2021object} have dedicated specific effort to solving just this problem, highlighting that it remains an outstanding challenge without simple off-the-shelf solutions. % (i.e., in comparison to the off-the-shelf nature of generating collision-free paths in environments modeled as 3D meshes). \asnote{maybe comment on why not using a coarse occupancy grid?}
As this was not the focus of our work, we used domain knowledge of the task and extra supervision to simplify path planning. Our approach was to create multiple Cartesian waypoints for the arm to reach along its path to the goal, each of which has a high likelihood of being collision-free (e.g., at a position high above the center of the table). 

\myparagraph{Additional ``offset'' waypoint pose in demonstrations} We also annotated an extra ``offset waypoint'' in the demonstrations, and used this extra waypoint annotation to solve for an offset relative to the inferred placement pose at test-time. This offset pose could be reached more easily without using fine-grained collision detection. Once at this offset pose, the final placement pose could be achieved by moving the end-effector in a straight-line task-space path. For example, in the ``mug on rack'' task, this was a pose where the mug's handle was aligned with the rack's peg, but at a position just in front of the tip of the peg. We performed this offset pose annotation in the demonstrations by recording an additional end-effector pose before moving the gripper to the final placement pose ($\demopose_{\text{place}}$). We then solved for the relative transformation between this offset pose and the placement pose. Finally, at test-time, we solved for the world-frame offset pose using this same relative transformation and the recovered placing pose. % of the task-relevant part of $\object_{A}$. % to circumvent solving the full collision-free planning problem in its entirety and

\subsection{Limiting Assumptions, Resulting Implications, and Avenues for Future Work}

\subsubsection{} \textbf{Modeling assumption: We model rearrangement tasks as SE(3)-transformation(s) of rigid bodies. R-NDF does not deal directly with other constraints like collision avoidance and kinematics, and depends on a separate motion planner for path planning.}
As discussed in the subsection above, full collision-free planning was not the focus of this work, and we therefore used heuristic solutions, domain knowledge, and extra real-world supervision for the purpose of simplifying path planning to avoid issues with additional constraints. 
We also describe in Section~\ref{sect:collision-avoidance} how to add extra costs that take more problem constraints into account.
Finally, note that the design choice of predicting relative transformations representing goal configurations allows our system to generalize much more easily and efficiently than it would have if we tried to learn full trajectory generators or closed-loop policies.

\subsubsection{} \textbf{Input/Task assumption: R-NDF performs category-level manipulation with known categories, and depends on the availability of an offline dataset of 3D shapes from each category.}
Assuming a known category does limit the ability to directly apply our method to brand new objects in unseen categories. However, this assumption is also what helps support generalization to new shape instances, as the model learns to associate the way different shapes are similar to each other. Several ``category-level'' manipulation systems and dense correspondence models have been proposed in the past~\cite{gao2019kpam, gao2021kpam, florence2018dense, yen2022nerfsupervision, duggal2022tars3D, deng2021deformed}, each with the reasoning that it's useful to enable programming a specific robot skill, or learning a category-specific concept, that works across all instances from a category. Obtaining a module that ignores global features of the object and focuses on local parts that may be shared across categories is an exciting direction for future work.

Depending on an offline set of 3D objects is another limitation. Many of the advancements in 3D vision and graphics have also used ground truth shapes for training~\cite{mescheder2019occupancy,peng2020convolutional,park2019deepsdf}, but it would be ideal if we could obtain a system with similar properties that does not depend on the availability of ground truth 3D shapes. We leave this for future work to address.

\subsubsection{} \textbf{Input assumption: R-NDFs use instance-segmented point clouds with known identity (i.e., the system knows which point cloud corresponds to $\object_{A}$ and $\object_{B}$).}
As the focus of this work is to relax the “static secondary/environmental object” assumption from the original NDF work, and expand the scope of achievable rearrangement tasks, we directly inherited the assumption that point clouds have been accurately segmented from the background. Note that significant progress has been made on instance/semantic segmentation~\cite{he2017mask,xiang2020learning,back2022unseen,xie2021unseen} with other robotic systems deploying these as components to their overall pipeline~\cite{mousavian2019graspnet,danielczuk2021object}. Based on this trend, it’s not unreasonable to assume access to an off-the-shelf module that provides segmented point clouds. However, it’s a fair concern that today, these systems are not robust enough to be entirely ``plug and play'' without substantial engineering effort. This is especially true in new environments that cause a distribution shift from the data they were trained on, which can negatively affect the downstream task performance.

Changing the underlying NDF formulation to reduce the dependence on performant off-the-shelf perception modules also has the potential to expand the system's capabilities. An analogous progression in 6-DoF grasp generation has occurred and shown meaningful improvements in overall system capability (i.e,. see~\cite{mousavian2019graspnet} and ~\cite{sundermeyer2021contact} where ~\cite{sundermeyer2021contact} does not require segmentation). However, this improvement is roughly orthogonal to our proposed approach. 

Recent works on using different neural network encoders for 3D data that use local features have shown promising results in not requiring accurate segmentation~\cite{ryu2022equivariant,chatzipantazis2022se,chen20223d}. The implications of global vs. local encoding on generalization and robustness to diverse geometries have been discussed at length in various recent works on neural fields~\cite{peng2020convolutional,jiang2020local,chibane2020implicit,chabra2020deep,neuralfieldsvisualcomputing2022}. In principle, transferring such approaches to our setting within the proposed R-NDF framework ought to provide similar benefits in the multi-object rearrangement tasks we consider. 

\subsubsection{} \textbf{General limitation: We use NDF as the core component of the framework. If the descriptors learned in NDF pretraining don't work well, then the downstream R-NDF framework also won't work well. What if they fail on more difficult shapes?}
It’s a concern that if the underlying NDF models have not learned meaningful descriptors that encode correspondence in a correct/useful way, our approach will inherit these problems and suffer in performance.
This could potentially occur on more difficult shape categories or with more difficult tests of generalization.

Recent approaches in neural implicit modeling include components for learning deformation fields with explicit correction terms~\cite{deng2021deformed} and latent spaces with higher-dimension~\cite{duggal2022tars3D} to recover cross-instance correspondence for topologically varying shapes. These ideas could potentially be incorporated into future versions of the NDF to support improved performance more challenging objects. Investigating such changes to the underlying NDF setup was beyond the scope of this work, but it’s an important consideration for scaling the approach to enable more difficult object categories.

\subsubsection{} \textbf{General limitation: This work only shows results in empty scenes with minimal clutter, no distractor objects, and multiple cameras to help retrieve a relatively complete point cloud.}
We separated out the issue of handling point clouds with heavy occlusions from the goal of relaxing the ``fixed placement object'' assumption from~\cite{simeonov2022neural}. However, as discussed in Section~\ref{sect:partial-pointclouds} and shown in Figure~\ref{fig:partial-pointcloud}, the NDFs can work with partial point clouds under some specific settings. First, similar occlusion patterns should be included in the training distribution for 3D reconstruction pretraining. Second, the task-relevant part of the object that was used for obtaining the target pose descriptor should not be entirely out of view. Handling partial point clouds with full generality may require incorporating active perception that searches in camera pose space, and we decided to avoid complicating the system by factoring in this set of considerations. 

New approaches to setting up the underlying NDF model (i.e., with point cloud encoders that operate on more local features) also have the potential to improve upon this issue. See the above paragraph regarding the quality of the underlying NDF models for further commentary on the potential implications of global vs. local feature encoders. 

\subsubsection{} \textbf{Modeling assumption: We model relational rearrangement via pair-wise relations. Can this be extended to general N-body rearrangement tasks?}
R-NDF can be applied to rearrangement tasks with more than two objects. Depending on the nature of the multi-object task specification, there are different approaches available, including sequential localization and composing multiple NDF descriptor distance terms in the optimization objective. See Section~\ref{sect:multi-object-rearrangement} for further discussion and examples for handling rearrangement with more than two objects. 

\subsubsection{} \textbf{Task limitation: We only show pick-and-place results. Can NDFs be used for tasks other than pick-and-place?}
NDFs can be used whenever it’s useful to localize a coordinate frame near a task-specific local part of an object. Inferring SE(3) transformations for pick-and-place with rigid objects is one instantiation of this. However, there maybe other settings where this is useful. For example, in multi-finger manipulation, it may be useful to encode the pose of each fingertip relative to an object with a per-finger query point set and corresponding fingertip-pose descriptor.

Another point to consider is that other tasks of interest to the community may potentially be recast as a form of pick-and-place. For instance, tool use consists of grasping an object, and then using some distal part of the grasped object to interact with the external environment. The geometric part of this “distal object part / environment” interaction could be solved in a way that is directly analogous to our ``placing pose'' inference. However, one fundamental limitation is that NDFs are a purely geometric representation. It would be interesting to see how to bring in dynamic/material/physical properties like mass, stiffness, friction, etc. and solve more dynamic tasks.

\subsubsection{} \textbf{Implications of these assumptions and limitations: There are many moving parts to the overall R-NDF system and this creates multiple potential sources of error. Which considerations impact the performance the most?} 
There can be multiple sources of error in NDF descriptor matching. A few common error sources, roughly in order of their impact on performance, include:
\begin{itemize}
    \item Out-of-distribution/severe point cloud noise. This can come from depth sensor noise and bad segmentation. For example, some of the real-world racks we experimented with were quite thin, shiny, and reflective, causing the depth image to have holes. The background/table was also sometimes not cleanly removed from the point cloud, which left regions containing outliers that looked different from the data the PointNet encoder was trained on. A similar issue can occur if the cameras are not accurately calibrated relative to each other.
    \item Inaccuracies in the demonstrations. For example, in the real world, if the object moves while in the gripper, or the placement object is accidentally bumped, the final point clouds we record might be in a different location than the true real-world objects.
    \item NDFs that pick up on task-irrelevant features. For example, a common failure mode for hanging the mug on the rack is to rotate the mug to almost exactly the correct orientation, but use the wrong rotation about the cylindrical axis. In this case, the NDF matches the demonstrations via a the side and the opening, rather than the handle. Similarly, a somewhat common failure mode for “bottle in container” is to place the bottle upside down, which can occur when the top and the bottom both have a locally planar geometry.
\end{itemize}

\myparagraph{What are the most important steps to help mitigate these errors?} 
To complement the discussion on multiple sources of error above, we list a set of key factors to pay particular attention to for reducing the likelihood of errors occurring:
\begin{itemize}
    \item As discussed above, out-of-distribution point clouds are a common source of error that can cause failure in descriptor matching. The best remedy for this is to train the NDF on diverse point clouds that cover the distribution that is likely to be encountered at test time. This can come from diverse random object scaling, point cloud masking, camera viewpoints, and other forms of 3D data augmentation.
    \item To complement the above point, making sure that the underlying implicit representation can fit the training data well for 3D reconstruction is important to ensure that the descriptors encode something meaningful. A useful debugging step when descriptor matching has errors is to examine the reconstruction and make sure it looks reasonable.
    \item We have found performance to be most consistent when using a handful ($\sim$10) of demonstrations, where the demonstrations themselves are somewhat diverse. They can be diverse in the shapes that are used and the pose of the objects in each demonstration. Using more demonstrations provides a better opportunity for the average target descriptor to accurately capture the task-relevant geometry, based on what is most saliently shared across the set of diverse demos. The issue of using a single demonstration (or a set of demonstrations that fail to disambiguate the task-relevant object parts) is discussed in Section 4.1 in the paper.
    \item With the current version of the framework, accurate segmentation, and point cloud outlier removal are important. We notice a meaningful improvement when taking serious care to completely remove everything in the background/nearby environment, and leave just the object in the point cloud. 
    \item It is important to run the NDF optimization from multiple diverse initializations. Because the optimization objective is non-convex, there exist many local minima. We have somewhat reduced the effects of this by moving toward a DeepSDF-based NDF (rather than the original OccNet-based NDF, see Section~\ref{sect:ndf-training}), but successfully obtaining a global optimum is still subject to running pose optimization from multiple initial values.
    \item Real-world execution requires collision-free motion planning to work properly. The whole execution is prone to fail if the placement object is accidentally bumped by the grasped object during placing. Providing intermediate waypoints to reach during the path execution that are conservatively far away from any potential collisions increases the likelihood of success. We also assume the object doesn’t move in the grasp while it’s moving to the placing pose. More robustness could be achieved by tracking the motion of the object in the grasp to take any unanticipated motion into account and adjust the final placing pose.
\end{itemize}

\vfill

\color{black}
\pagebreak
\section{Model Architecture Diagrams}
\label{sect:model-arch-diagrams}

\begin{figure}[H]

\begin{minipage}[b]{0.45\linewidth}
\centering
\begin{tabular}{c}
    \toprule
    \toprule
    VN-LinearLeakyReLU 128 \\
    \midrule
    Linear 128 \\
    \midrule
    VN-ResnetBlock FC 128 \\
    \midrule
    VN-ResnetBlock FC 128 \\
    \midrule
    VN-ResnetBlock FC 128 \\
    \midrule
    VN-ResnetBlock FC 128 \\
    \midrule
    VN-ResnetBlock FC 128 \\
    \midrule
    Global Mean Pooling \\ 
    \midrule
    VN-LinearLeakyReLU $\rightarrow$ 256 \\ 
    \bottomrule
\end{tabular}
\captionof{table}{Vector Neuron point cloud encoder architecture}
\label{fig:vnn-encoder}
\end{minipage}%
\hfill
\begin{minipage}[b]{0.45\linewidth}
\centering
\begin{tabular}{c}
    \toprule
    \toprule
    ResnetBlock FC 128 \\
    \midrule
    ResnetBlock FC 128 \\
    \midrule
    ResnetBlock FC 128 \\
    \midrule
    ResnetBlock FC 128 \\
    \midrule
    ResnetBlock FC 128 \\
    \midrule
    Linear $\rightarrow$ 1 \\
    \bottomrule
\end{tabular}
\captionof{table}{Signed-distance function decoder architecture}
\label{fig:vnn-decoder}
\end{minipage}\hfill

\end{figure}

\begin{figure}[H]
\begin{minipage}[b]{0.45\linewidth}
\centering
\begin{tabular}{c}
    \toprule
    \toprule
    Dense $\rightarrow$ 128 \\
    \midrule
    Dense $\rightarrow$  128 \\
    \midrule
    Dense $\rightarrow$ 6 \\
    \bottomrule
\end{tabular}
\captionof{table}{Architecture of EBM capturing relations.}
\label{tbl:ebm}
\end{minipage}
\hfill
\begin{minipage}[b]{0.45\linewidth}
\centering
\begin{tabular}{c}
    \toprule
    \toprule
    Dense $\rightarrow$ 256 \\
    \midrule
    Dense $\rightarrow$  256 \\
    \midrule
    Dense $\rightarrow$ 6 \\
    \bottomrule
\end{tabular}
\captionof{table}{Architecture of pose regression baseline.}
\label{tbl:baseline_pose_regress}
\end{minipage}%
\end{figure}

\clearpage

\end{document}